\documentclass[preprint,12pt]{elsarticle}
\usepackage[pdfborder={0 0 0}]{hyperref}
\usepackage{amsthm,amsmath,amsfonts,latexsym,amssymb}
\usepackage{graphics,fancyhdr,graphicx}
\usepackage[ruled,linesnumbered]{algorithm2e}
\usepackage[margin=1in]{geometry}
\usepackage{multirow,booktabs}
\usepackage[table]{xcolor}
\usepackage{listings}
\usepackage{tabularx}
\usepackage{placeins}
\usepackage{comment}
\usepackage{lscape}
\usepackage{bm}
\usepackage{hyperref}
\usepackage{mathrsfs}
\usepackage{arydshln}
\usepackage{appendix}
\usepackage{soul}
\definecolor{custom-blue}{RGB}{3,69,173}
\hypersetup{
    colorlinks = true,
    urlcolor   = blue,
    citecolor  = custom-blue,
}
\biboptions{numbers,sort&compress}
\definecolor{listinggray}{gray}{0.9}
\definecolor{lbcolor}{rgb}{0.9,0.9,0.9}
\definecolor{Darkgreen}{RGB}{0,100,0}
\usepackage[font=small,labelfont=bf, textfont=it]{caption}

\usepackage{float}
\usepackage{booktabs}
\usepackage{array}
\usepackage{multirow}
\usepackage{tabularray}
\usepackage{hhline}
\usepackage{algpseudocode}
\usepackage{mathtools}
\usepackage{subcaption}
\usepackage{enumitem}
\usepackage{lipsum}
\usepackage[most]{tcolorbox}

\makeatletter
\def\ps@pprintTitle{%
 \let\@oddhead\@empty
 \let\@evenhead\@empty
 \def\@oddfoot{}%
 \let\@evenfoot\@oddfoot}
\makeatother

\newtcolorbox{textbox}[1]{
    title={\hspace*{0mm}#1},
}

\begin{document}
\abovedisplayskip=6.0pt
\belowdisplayskip=6.0pt

% \linenumbers

\newcommand{\Review}[2][blue]{{\textcolor{#1}{#2}}}

\begin{frontmatter}

\title{Toward Autonomous Engineering Design: A Knowledge-Guided Multi-Agent Framework}

\author[1]{Varun Kumar}
\ead{varun_kumar2@brown.edu}
\author[2]{George Em Karniadakis\corref{cor1}}
\ead{george_karniadakis@brown.edu}

\address[1]{School of Engineering, Brown University}
\address[2]{Division of Applied Mathematics, Brown University}

\cortext[cor1]{Corresponding author.}

\begin{abstract}
\noindent
The engineering design process often requires expertise from multiple domains, leading to complex collaborations and iterative refinements. Traditional approaches can be resource-intensive and prone to inefficiencies. To address this, we present a multi-agent reasoning framework that formalizes the engineering design process through structured, iterative design and review loops. The framework introduces specialized, knowledge-based AI agents that collaborate to generate, evaluate, and refine design candidates efficiently. 
As a demonstration, we apply this framework to the design of four-digit NACA airfoils for maximizing lift-to-drag ratio. The framework consists of three key agents: a \emph{Graph Ontologist}, a \emph{Design Engineer}, and a \emph{Systems Engineer}. The \emph{Graph Ontologist}, powered by a Large Language Model constructs two domain-specific knowledge graphs using airfoil development literature. The \emph{Systems Engineer}, guided by the \emph{Manager's} (human) input, defines technical requirements that inform the design process. The \emph{Design Engineer} utilizes these requirements, alongside the design knowledge graph and design-specific tools, to produce candidate airfoil designs. The \emph{Systems Engineer} then reviews these designs using its own knowledge graph, providing both qualitative and quantitative feedback for iterative design improvements. A continuous feedback loop exists between the \emph{Design Engineer} and \emph{Systems Engineer} until the \emph{Manager} validates a design, at which point the iterative phase ends. The valid design is subsequently optimized to achieve performance objectives. 
In summary, this framework demonstrates a path toward improving efficiency and quality in the engineering design process by leveraging collaborative AI agents equipped with domain-specific knowledge representation.

 \end{abstract}
 
\end{frontmatter}

\section{Introduction}
\label{sec:intro}
The engineering design process serves as the cornerstone of innovation in diverse disciplines. At its core lies an iterative cycle - a structured approach to problem-solving that emphasizes refinement and improvement through repeated cycles of design space exploration, analysis, evaluation, review, and selection. This iterative nature arises due to the complexity of real-world engineering challenges, where final solutions often emerge after multiple rounds of design reviews and requirement clarifications. The process of engineering design, therefore, is not a linear sequence of events but rather an intertwined set of activities, which are often revisited and reordered as the problem understanding deepens. Traditionally, the design process is segmented into three key phrases: conceptual design, embodiment design, and detailed design \cite{PahlBeitz}. The conceptual design phase involves initial exploration of potential solutions, focusing on defining the problem scope, requirement elicitation, generation of several design concepts and their evaluation against requirements. Often, this phase is characterized by a high degree of uncertainty and repetitions, since the goal is to identify promising directions for further detailed investigation. 

In addition to its iterative nature, the design process is also a collaborative endeavor, particularly in the context of complex product development. Interdisciplinary teams, comprising of experts and personnel from diverse specialties are essential for bringing a holistic perspective in this process. Each discipline contributes their unique expertise and knowledge-base, leading to more robust and innovative solutions. This collaborative environment necessitates frequent communication and coordination, often facilitated by specialized design tools and processes. Decision making is an essential step in design and is typically performed by domain experts whose expertise is built with years of practice. This domain knowledge is often tacit, and is difficult to articulate or codify \cite{POLANYI_tacit}. Such knowledge encompasses not only technical skills but also an understanding of the best practices, potential pitfalls, and subtle nuances of design trade-offs. This accumulated expertise is immensely valuable for ensuring process efficiency, quality, and risk management. However, this reliance on individual expertise presents a significant challenge: the potential loss of critical knowledge during personnel transitions \cite{Argote_org_learning}. The departure of experienced engineers can create a void in institutional memory, hindering future design efforts and potentially decline in product quality \cite{personal_knowledge_management}.

The emergence of large language models (LLMs) marks a transformative shift in the traditional engineering design landscape. These models, trained on large collections of text and code datasets, have shown an unprecedented ability to understand, generate, and perform complex tasks when provided with well-designed instructions \cite{instructgpt}. The proliferation of LLMs in engineering practice has the potential to reshape the entire design process. LLMs can offer compelling advantages as assistant tools in guiding users through complex, iterative processes that characterize engineering design. In future, one can visualize LLM frameworks capable of generating design concepts, explaining the rationale behind design choices, identifying potential risks and remedies. Such interactive guidance can be particularly valuable for novice engineers, accelerating their learning curve and fostering a deeper understanding of design principles and engineering tools. 

Despite their remarkable capabilities, the application of LLMs for engineering tasks is not without limitations. A particularly significant challenge lies in achieving domain specific knowledge and guidance. Although LLMs excel at general language understanding, they often struggle to provide the deep and specialized knowledge required for complex engineering tasks that frequently require intricate trade-offs and nuanced considerations \cite{LLM_engg_tasks}. Furthermore, the backbone of engineering practice relies on specialized engineering tools—computational fluid dynamics (CFD) solvers, finite element analysis (FEA) software, CAD modeling packages—that have been rigorously developed and validated over many years. Engineers depend on these tools to perform critical calculations, simulations, and analyses. LLMs typically lack explicit knowledge of these tools, making it difficult to accomplish complex engineering tasks solely through LLM-driven interactions without human intervention \cite{toolformer, easytool}. The seamless integration of LLMs with existing engineering workflows and tools remains a significant hurdle to overcome \cite{chiarello2024generative}.

This work addresses the challenges and opportunities presented by the confluence of iterative design processes and the rise of LLMs by introducing a novel multi-agent framework for use in iterative design and review processes, particularly within the {\em conceptual} design phase. We focus on the pedagogical problem of designing 4-digit NACA airfoils for maximizing performance—a common and accessible exemplar problem in aerospace engineering—to demonstrate the capabilities of our multi-agent framework. The framework comprises three specialized AI agents, each assigned a distinct purpose and contributing to the overall design process: a \emph{Graph Ontologist}, a \emph{Design Engineer}, and a \emph{Systems Engineer}. The Graph Ontologist acts as a knowledge curator, leveraging LLMs to extract and structure information from existing literature pertaining to airfoil design and development, helping to create two distinct knowledge graphs, specific for each engineering agent. These two graphs serve as source of background information, accessible to the designated agents within the framework. The Design Engineer agent utilizes the designer's knowledge graph and design-specific tools to generate and refine airfoil designs, guided by the knowledge contained within the graph. The Systems Engineer agent is tasked with translating the human manager's request into formal technical requirements, and also reviews the designs generated by the Design Engineer agent, providing feedback for improvements.

\begin{figure}[h]
    \centering
    \includegraphics[width=1\linewidth]{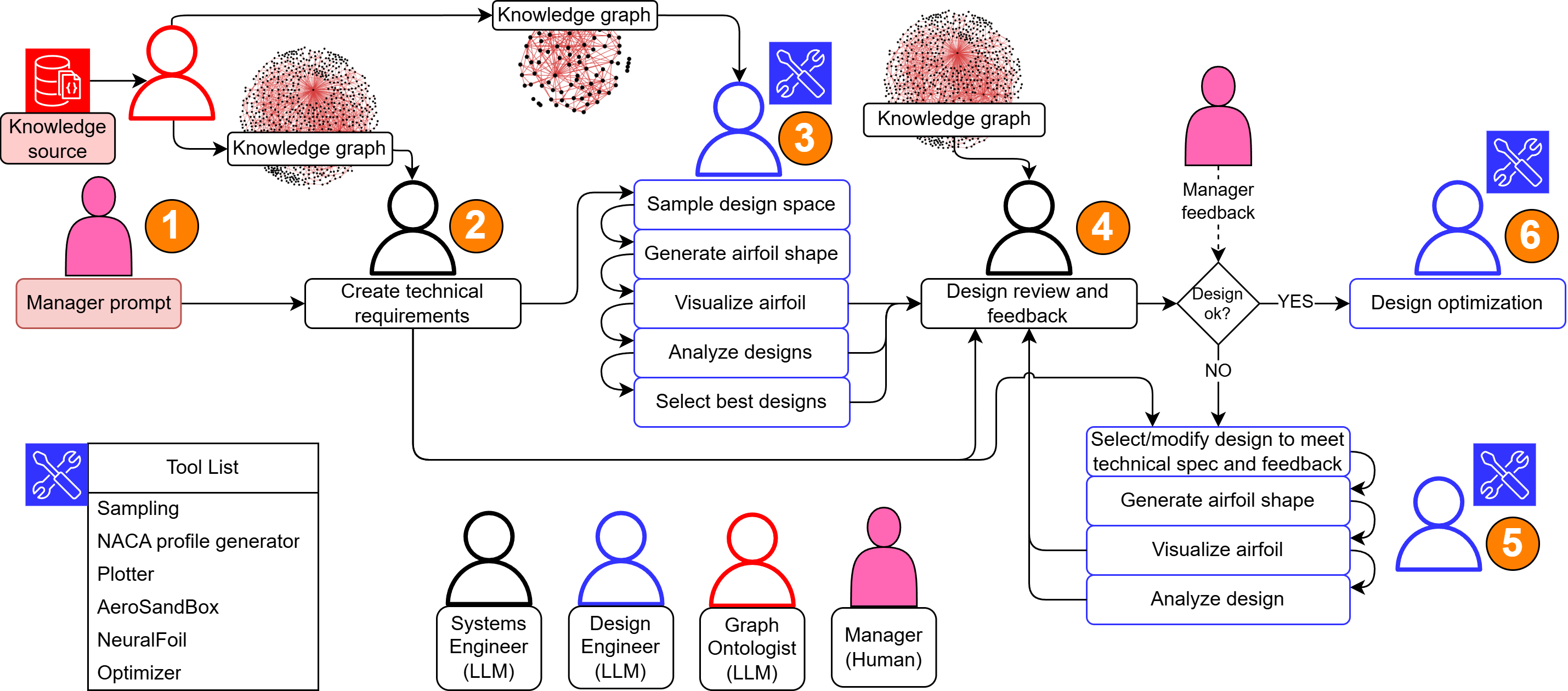}
    \caption{Schematic for LLM-based multi-agent workflow, consisting of three LLM agents namely \emph{Graph Ontologist}, \emph{Design Engineer}, \emph{Systems Engineer}, each assigned a designated set of tasks in this process. The Design and Systems Engineer agents are provided with their respective knowledge graphs tailored specific to their roles. The Design Engineer agent possesses a set of engineering tools to help him accomplish his tasks. The Systems Engineer agent utilizes a multi-modal language model for reviewing airfoil shapes generated by the Design Engineer and provides design improvement suggestions. The Design Engineer then acts on the feedback, updates the design, and sends it for another round of review process. This process also allows for a Manager (human user in this case) to provide feedback for design improvement. The design-review cycle terminates once the Manager determines the design to be a suitable candidate for further optimization. Details for each component and steps in this workflow are discussed later in Section \ref{sec:methodology}.}
    \label{fig:workflow}
\end{figure}

The agents accomplish the design process through an iterative feedback loop. The Design Engineer agent proposes a design, the Systems Engineer agent evaluates it against the defined requirements and provides feedback, and the Design Engineer agent iterates on the design based on that feedback. This cycle repeats until the Manager (human) deems a design as acceptable in conjunction with the Systems Engineer qualifying the design as `valid' against the requirement set. This approach aims to combine the knowledge-curation capabilities of LLMs with the rigor of established engineering practices, creating a powerful tool for iterative design and review. The complete workflow for this framework is shown in Figure \ref{fig:workflow}.

\section{Background and Related Works}
\label{sec:Lit_review}
Multi-Agent Systems (MAS) have long been proposed to support engineering design through distributed decision-making, concurrent workflows, and interdisciplinary coordination. Early work in the 1990s emphasized knowledge-based and protocol-driven agents for concurrent engineering. Foundational systems such as PACT demonstrated ontology-based coordination among agents for cooperative design refinement \cite{cutkosky1993pact}. Similar MAS approaches were applied to cable harness design and distributed design environments, highlighting benefits of problem decomposition and message-based coordination but also exposing limitations due to hand-crafted rules and communication overhead \cite{park1994agent, shen1995dide, shen1996experimental}. Subsequent research explored alternative coordination mechanisms, including market-based agent marketplaces for concurrent, set-based design \cite{parunak1997marketplace}. In conceptual design, agent-based configuration synthesis systems such as A-Design employed specialized agents to generate, evaluate, and refine electro-mechanical concepts using Pareto-based selection \cite{campbell1998agent, campbell1999design}. However, these approaches assumed well-defined objectives, which are often unavailable in early design stages.

With the maturation of CAD/CAE and PLM infrastructures, MAS research shifted toward distributed collaboration and interoperability. Agent frameworks supported manufacturability, cost, and lifecycle analysis by coordinating heterogeneous tools through facilitator agents and shared ontologies \cite{Sun_2000, Sun_2001, lees2001framework, zhao2001clover}. Domain-specific systems addressed structural, process, and assembly design, as well as knowledge management and process coordination, but incurred high knowledge-engineering costs \cite{anumba2001multi, chao2002agent, zha2002knowledge, huang2004multi, madhusudan2005agent}. More recent work integrates MAS into CAD-adjacent pipelines, enabling multi-level 3D collaboration, sustainability-aware design, and automated analysis–optimization loops \cite{chu2009multi, dostatni2016multi, gembarski2020agent}. Knowledge-based and Belief-Desire-Intention (BDI)-style agents have further enabled explainable design evaluation and digital design reviews, reinforcing MAS potential while highlighting ongoing challenges in scalable knowledge maintenance and geometric reasoning \cite{plappert2021knowledge, plappert2022development, plappert2023scalable}.

More recently, the rapid advancement of large language models (LLMs) has created a new research direction - development and use of LLM-powered MAS within engineering contexts. Before the rise of MAS frameworks, several studies have demonstrated the value of individual LLM assistants in engineering workflows \cite{mycrunchgpt, gomez2024large, makatura2024can_part1, makatura2024can_part2, picard2025concept}. Recent research ideology, however, prescribes MAS frameworks as being more suited for handling end-to-end design workflows, since different agent skill sets is desired at different stages of design. Early explorations of LLM‑driven multi‑agent systems such as DesignGPT \cite{ding2023designgpt} demonstrated a design workflow targeted at using generative AI image generation for generic product design. However, this framework does not address a specific engineering problem requiring domain knowledge and, hence, is prone to hallucinations. \cite{texttodesign} and \cite{ideatoCAD} propose MAS frameworks wherein user requirements, expressed as text or sketches, are translated into CAD models via coordinated LLM agents responsible for requirements engineering, geometric modeling, and vision-based verification. 

\cite{markus_beuhler} introduced a LLM assisted agent to construct and query a large ontological graph derived from 1000 scientific papers published in the area of biological material design. \cite{sciagents} further extended this to MAS, to autonomously generate and refine research hypothesis for new material discovery. \cite{ferrag2025llm} reviewed the path from LLM reasoning to autonomous agents, providing general design of agent tool-use, memory, and multi-agent coordination. \cite{obieke2025framework} proposed the framework for AI collaboration in engineering design (AICED), where four LLM agents accomplish tasks associated with market research, documentation, design, and manufacturability using required design knowledge from professionals. In \cite{elrefaie2025ai} a MAS framework was introduced to automatically perform tasks such as conceptual sketching, shape retrieval, and CFD meshing by integrating a set of agents within the automotive design workflow. \cite{zhang2025idesigngpt} advanced LLM agentic workflows tailored to engineering design, by automating requirement gathering and evaluation criteria using design tools such as TRIZ \cite{TRIZ_intro} and QFD \cite{QFD}. This framework utilized design loops to evaluate and refine conceptual designs for the autonomous emergency response aircraft problem \cite{GoAero}.

Mechanical Engineering Design Agents (MEDA) \cite{panta2025meda} introduced MAS for parametric CAD model creation, mapping design intents into CAD features via collaborating agents. In \cite{cost_aware_multigen}, a cost-aware MAS was proposed for design space exploration for search under expensive black-box objectives, to model the sequential-decision making process of a design team. This method uses a combination of Bayesian Optimization and Reinforcement Learning to distribute the task of design space exploration between multiple agents. \cite{ghasemi2025vision} utilized vision-language models in an Generator-Critic framework for concept generation and evaluation where the Generator generates initial concept which is refined iteratively based on Critic feedback. The Critic provides both qualitative and quantitative feedback on design suitability. In \cite{massoudi2025agentic}, a multi-agent framework is used for conceptual design of solar-powered water filtration system, where dedicated agents perform functional decomposition, design generation, evaluation and grading in an iterative cycle. We note, however, that this framework relies on existing knowledge base of LLM agents and does not use human-in-loop ideology. \cite{wang2025llm} propose a multi-agent framework for designing autonomous water quality monitoring system, which involves agents representing mechanical designer, simulation, and electronics designer. The mechanical designer uses a set of code-based tools for ideation while the electronics designer uses knowledge acquired from technical data sheets for proposing suitable electronic components. At a higher level, \cite{jiang2025intelligent} articulates a paradigm shift toward adopting agentic AI framework for automation of engineering processes.

Based on a review of existing literature, we observe that integrating MAS into engineering design workflows presents substantial challenges due to the inherently collaborative, iterative, and domain-knowledge intensive nature of engineering practice. Traditional design relies on coordinated expertise from multiple specialists—such as structural, materials, and manufacturing engineers—whose reasoning must either be encoded in large knowledge bases or learned from extensive datasets, both of which are resource-intensive \cite{multiagent_design_isssues, AI_modern_approach}. Effective MAS deployment further requires deep domain knowledge, including design factors, manufacturing constraints, and regulatory requirements, which is difficult to represent in forms suitable for effective autonomous agent reasoning. Additionally, an ideal MAS system should be able to automatically adjust design parameters, rerun analysis, and flag potential design issues based on a specific set of knowledge. However, current systems often lack the sophisticated adaptive decision-making capabilities required for such nuanced problem-solving \cite{llm_decision_making}.  While multi-agent frameworks offer a promising avenue for encoding and managing such domain expertise, enabling them to interact with existing engineering tools and adapt to iterative design processes remains a research opportunity. Hence, the key objective of our work includes the following:
\begin{itemize}
    \item Utilize LLM-agents equipped with existing domain-specific knowledge in an `Generator-Critic' setup for fulfilling specific engineering roles.
    \item Integrate MAS into an iterative design framework where outcomes from one LLM agent assist another agent with its decision making and task completion.
    \item Associate dedicated engineering tools with agents to enable successful task completion.
    \item Allow human-in-loop interaction to ensure necessary check and balances in the process.
\end{itemize}

\section{Problem Setting} 
\label{sec:prob_def}
The design of NACA airfoil profiles has historically played a crucial role in the advancement of aerodynamics in aviation \cite{orders_of_mag_NACA}. The National Advisory Committee for Aeronautics (NACA) developed families of airfoils and categorized them based on their geometric and aerodynamic properties, with the 4- and 5-digit series being widely used. The significance of NACA airfoils predominantly lies in their predictable aerodynamic performances, making them instrumental in both academic research and practical applications \cite{wing_section_theory_abbot}. Their predictable behavior simplifies the analysis of how airfoil shape affects lift, drag, and overall aerodynamic efficiency, and serves as a pedagogical example for demonstrating our multi-agent design workflow.

The airfoil shape is defined by parameters such as maximum camber, maximum thickness, and maximum camber position. These parameters significantly influence the aerodynamic performance of the airfoil. Small variations in these parameters can lead to substantial changes in the pressure distribution around the airfoil, and thus, its overall aerodynamic performance \cite{wing_section_theory_abbot_appendix}. For instance, increasing the maximum camber generally increases lift but also increases drag, while repositioning the maximum camber affects the pressure gradients and can influence stall characteristics. Understanding and precisely controlling these relationships is crucial for effective airfoil design. The four-digit NACA equation is foundational in the generation of these airfoil shapes \cite{wing_section_theory_abbot}, and is given by Equation \ref{eqn:NACA_profile}:

\begin{align}
y_t = 5t(0.2969 \sqrt{x} - 0.1260x - 0.3516x^2 + 0.2843x^3 - 0.1015x^4), \label{eqn:NACA_profile}
\end{align}
where $y_t$ is the half thickness at a given position $x$ along the chord, and $t$ represents the maximum thickness as a fraction of the chord. In this work, we consider the design space defined by airfoil parameters max-camber ($c_{max}$), max-camber location ($c_{loc}$), and max-thickness ($t_{max}$)
\begin{align}
    \mathcal{D}:=[0.01, 0.095] \times [0.05, 0.9] \times [0.01, 0.40], \quad c_{max}, c_{loc}, t_{max} \in \mathcal{D}.
\end{align}

The central goal of this research is to design an efficient NACA airfoil profile, specifically targeting a maximum lift-to-drag ratio at  Mach number 0.6 and  Reynolds number  $5 \times 10^6$. This specific operating condition represents a typical test condition for airfoil design validation \cite{sewall1985wind}. Furthermore, for simplicity during the initial design phase, we assume a 0-degree angle of attack. This simplification allows us to focus on the fundamental shape selection and optimization steps without the complexities introduced by the angle of attack. The design problem is chosen for its simplicity and the abundance of readily available technical documentation, which will be essential for constructing knowledge graphs in subsequent stages of the research. This problem definition can be adapted based on managerial requirements, but in the current scope, we concentrate solely on the aforementioned conditions.  Furthermore, non-functional constraints will arise from the specific geometric parameters chosen for airfoil design. For example, excessively sharp leading edges, while potentially reducing drag, may be structurally impractical for manufacturing and operation. Such considerations are embedded as knowledge sources with our engineering agents, thereby enabling a holistic approach to design selection and evaluation process. 

Furthermore, to address this optimization challenge, advanced tools such as AeroSandBox \cite{aerosandbox_phd_thesis} and NeuralFoil \cite{neuralfoil} are utilized. AeroSandBox facilitates fluid simulations and aerodynamic analysis, while NeuralFoil offers machine learning capabilities to refine the design by anticipating aerodynamic characteristics of new airfoil shapes. These tools, when integrated, provide a robust framework for conducting design and analysis iteratively and efficiently. We discuss these tools in detail later in section \ref{sec:methodology}. It is essential to acknowledge that while this investigation is focused on a specific aerodynamic design problem, the methodology employed here is adaptable to other design scenarios.

\section{Methodology}
\label{sec:methodology}
\subsection{Knowledge Graphs for Engineering Agents}
To imbue our LLM agents with domain expertise and facilitate informed decision-making within the engineering design process, we leverage knowledge graphs (KGs) as a means of representing structured knowledge \cite{hogan_knowledge_graph}. KGs offer a powerful and efficient way to encode a large body of existing knowledge, going beyond simple text-based information retrieval \cite{yagoKG, dbpediaKG, freebaseKG}. The graph-based representation allows for the discovery of novel connections and ideas between concepts, even those that might appear distant based on human intuition but are ontologically linked. This ability to uncover hidden relationships is particularly valuable in complex engineering domains where tacit knowledge and interdependencies are prevalent. Our approach utilizes two distinct KGs, designed to represent differing levels of knowledge abstraction and, consequently, to simulate varying degrees of expertise within our LLM agents, mirroring the spectrum of expertise found in real-world engineering practice.

These KGs were generated using a dedicated Graph Ontologist agent \cite{markus_beuhler}, which operates according to a set of pre-defined instructions tailored to create agent-specific knowledge graphs. This controlled generation process ensures consistency and relevance of the knowledge encoded within each graph. The content for these graphs is derived from a comprehensive corpus of published reports and papers focusing on NACA airfoil design and development. This collection spans a considerable period, from the early years of NACA research in 1940s to more recent publications in 2010s, effectively capturing the evolution of knowledge and best practices accumulated over decades. This long-term perspective is crucial for mimicking the experience and understanding that an expert engineer develops over years of training and practical application. It avoids the pitfalls of relying solely on the latest information, ensuring that fundamental principles and historical context are also incorporated into the agents' knowledge base.

The use of KGs within our LLM agents provides a crucial foundation for ensuring that design choices are supported by consistent practices and prior findings. Without this structured knowledge, agents would be prone to generating inconsistent or suboptimal designs and responses. Furthermore, the KGs serve as a mechanism for transparency and explainability, allowing us to trace the origins of design recommendations and understand the reasoning behind agent decisions \cite{KG_explainable}. This capability is essential for building trust and facilitating human oversight within the design process. In the subsequent section, we will detail the processes involved in knowledge graph creation and how the Graph Ontologist agent constructs these resources.

\subsection{Graph Ontologist Agent}
\label{sec:graph_ontologist}
To facilitate the creation of specialized knowledge graphs for our Design and Systems Engineering agent, we employed a dedicated Graph Ontologist agent. This LLM agent was specifically designed to extract and structure knowledge from unstructured text, forming the foundation of our knowledge graph. The Graph Ontologist agent utilizes LLM capabilities to interpret context and infer semantic relationships. The corpus utilized for knowledge graph generation comprises of 50 scientific articles and reports spanning decades of NACA airfoil design and development. Initially, the raw data from the scientific papers was distilled through a sequence of preparatory steps. This included converting the PDF into a markup language format, followed by segmenting the text into manageable chunks. A chunk size of 3000 was chosen heuristically to represent semantically coherent units of information, hence allowing the Graph Ontologist agent to focus on localized knowledge extraction. Each text chunk was then processed by the Graph Ontologist agent to extract local knowledge graph triples, representing relationships between key concepts. These triples consisted of two nodes, each representing a significant concept from the extracted ontology, and an edge feature that defined the relationship between the two concepts. The triple set is defined as\footnote{Code for knowledge graph generation was obtained from \url{https://github.com/rahulnyk/knowledge_graph}}: `node1': “A concept from extracted ontology”, `node2': "A related concept from extracted ontology”, `edge': “relationship between the two concepts, node1 and node2 in one or two sentences”. This explicit definition guided the agent's extraction process, ensuring a consistent and interpretable knowledge representation. The procedure for KG generation is adopted from \cite{buehler_knowledge}.

\begin{figure}[t]
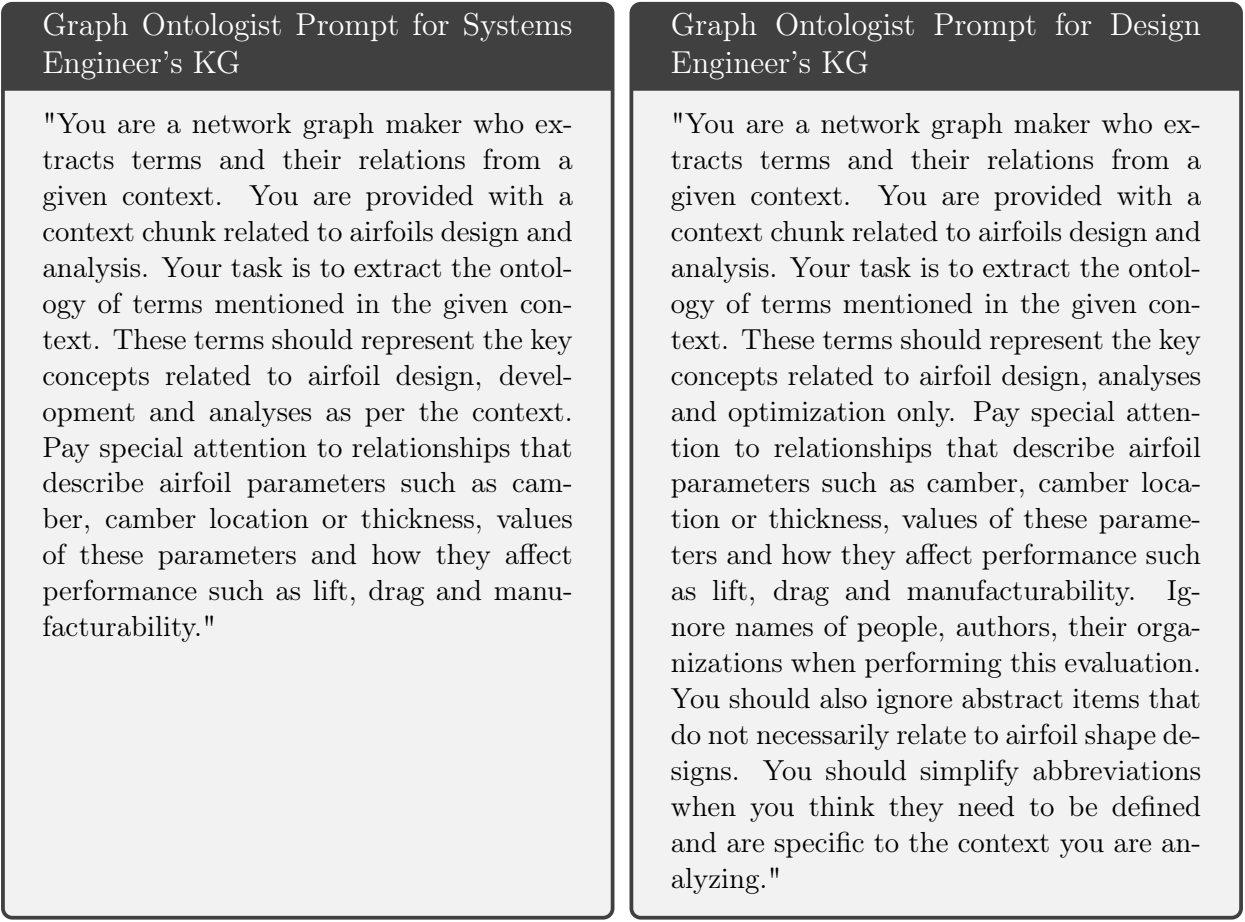

    \centering
    \begin{tcbraster}[raster columns=2,raster equal height,nobeforeafter,raster column skip=0.2cm]
    \begin{textbox}{Graph Ontologist Prompt for Systems Engineer's KG}
    \small
    "You are a network graph maker who extracts terms and their relations from a given context. You are provided with a context chunk related to airfoils design and analysis. Your task is to extract the ontology of terms mentioned in the given context. These terms should represent the key concepts related to airfoil design, development and analyses as per the context. Pay special attention to relationships that describe airfoil parameters such as camber, camber location or thickness, values of these parameters and how they affect performance such as lift, drag and manufacturability."
  \end{textbox}
  \begin{textbox}{Graph Ontologist Prompt for Design Engineer's KG}
  \small
    "You are a network graph maker who extracts terms and their relations from a given context. You are provided with a context chunk related to airfoils design and analysis. Your task is to extract the ontology of terms mentioned in the given context. These terms should represent the key concepts related to airfoil design, analyses and optimization only. Pay special attention to relationships that describe airfoil parameters such as camber, camber location or thickness, values of these parameters and how they affect performance such as lift, drag and manufacturability. Ignore names of people, authors, their organizations when performing this evaluation. You should also ignore abstract items that do not necessarily relate to airfoil shape designs. You should simplify abbreviations when you think they need to be defined and are specific to the context you are analyzing."
  \end{textbox}
\end{tcbraster}
    \caption{Prompts provided to Graph Ontologist for generating two different KGs for our engineering agents. For the Systems Engineer KG, a more generic set of goals leads to the creation of a larger domain of knowledge as compared to the Design Engineer's KG, which is more focused on aspects related to airfoil design.}
    \label{fig:graph_prompts}
\end{figure}

We employ two KGs in this study that differ significantly in scope and detail. The first KG, designed for the Systems Engineer agent, encompasses a broad range of aspects related to airfoil and aircraft development. This includes design considerations, fabrication processes, validation procedures, and experimental evidence. This KG represents the perspective of a seasoned systems engineer with extensive experience overseeing the development of aerodynamic systems across various product design contexts. The relationships within this graph emphasize the interdependencies between different stages of the design lifecycle and the impact of design decisions on overall system performance. The second KG, utilized by the Design Engineer agent, adopts a more focused perspective on specific design aspects influencing airfoil shapes and their subsequent aerodynamic performance. This graph primarily focuses on the relationships between design variables, their impact on key performance metrics, and typical values observed in practical applications. The focus is on creating a more granular knowledge base for the Design Engineer agent. To obtain these distinct KGs, two sets of prompts were provided to the Graph Ontologist agent as shown in Figure \ref{fig:graph_prompts}. The Mistral-NeMo model was utilized iteratively to extract these triples from the text chunks, leveraging its ability to understand complex sentence structures and identify subtle semantic relationships.

\begin{figure}[t]
  \centering
  % Left subfigure
  \begin{subfigure}[t]{0.48\textwidth}
    \centering
    \includegraphics[width=\textwidth]{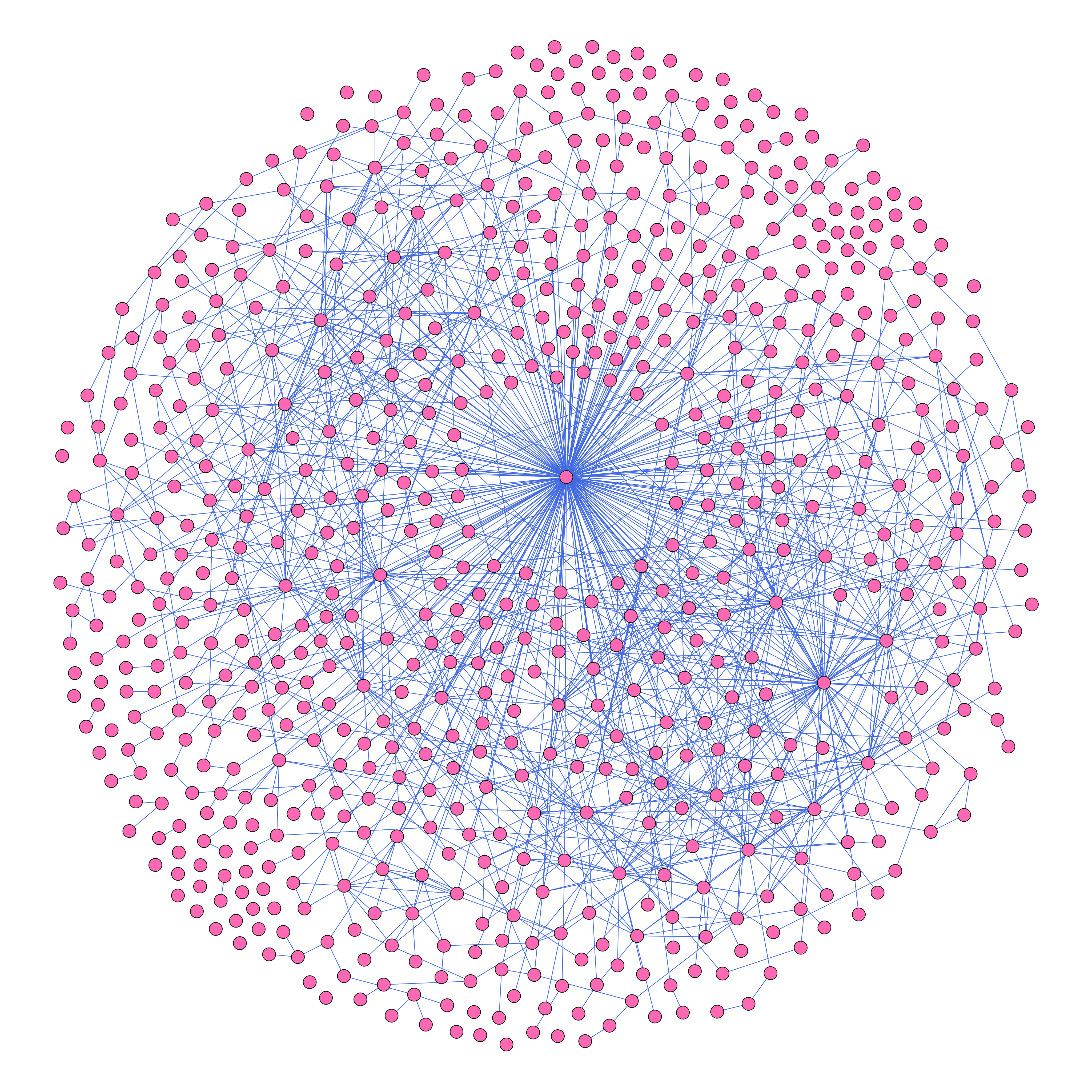}
    \caption{Knowledge graph used as knowledge base for Systems Engineer agent.}
    \label{fig:KG_sys_engg}
  \end{subfigure}
  \hfill
  % Right subfigure
  \begin{subfigure}[t]{0.48\textwidth}
    \centering
    \includegraphics[width=\textwidth]{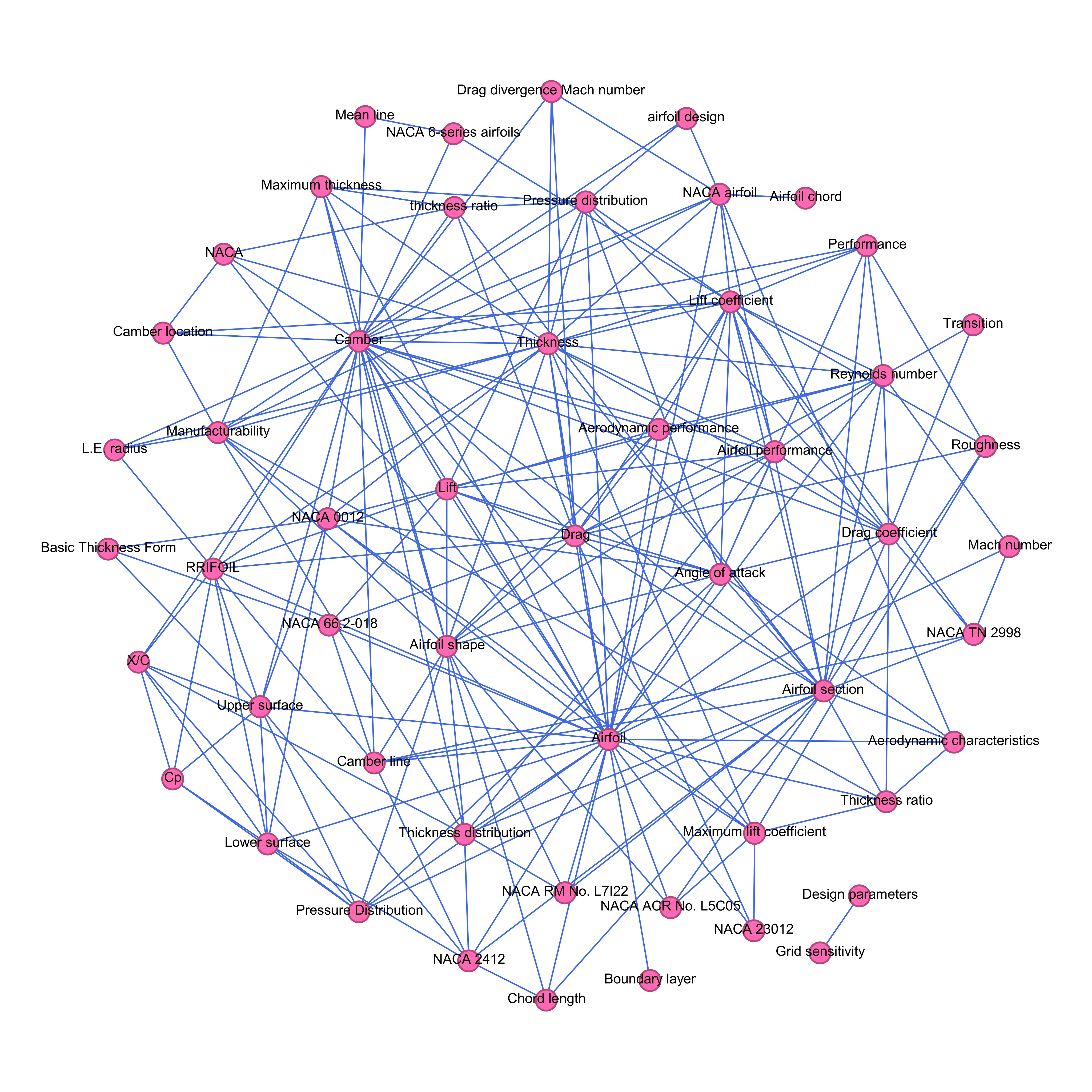}
    \caption{Knowledge graph used as knowledge base for Design Engineer agent.}
    \label{fig:KG_design_engg}
  \end{subfigure}
  \caption{Representation of Knowledge Graphs generated by the Graph Ontologist for the two engineering agents. Systems Engineer's KG contains $\approx 700$ nodes while the Design Engineer's KG contains $\approx 50$ nodes. Note that these graphs only contain nodes with a degree of 10 or more, the remaining nodes were filtered out for compactness of the knowledge base.}
  \label{fig:KGs}
\end{figure}

The individual local graphs generated from each text chunk were subsequently consolidated into a global knowledge graph. This process involved merging nodes representing similar concepts and eliminating ‘hanging’ nodes—those that were not connected to the rest of the graph, indicating potentially erroneous extraction. The merging of similar nodes was performed using a similarity metric based on node embeddings, ensuring that semantically related concepts were integrated into a unified representation. The edge weights were increased during the concatenation process, reflecting the cumulative evidence supporting the relationships between concepts. The resulting knowledge graphs for the two agents are visually represented in Figure \ref{fig:KGs}, illustrating the distribution of nodes and edges within each graph. The structural differences between the graphs, as reflected in their visual representation, highlight the impact of the different prompts used to guide the Graph Ontologist agent, demonstrating the adaptability and utility of this approach for creating specialized knowledge resources.

\subsection{Systems Engineer Agent}
\label{sec:systems_engg_agent}
The Systems Engineer agent embodies the responsibilities of a senior systems engineer, leveraging a comprehensive knowledge base and advanced reasoning capabilities to generate technical requirements, review designs, and offer recommendations for design improvements based on knowledge. The design of this agent is informed by established practices in systems engineering, where a systems engineer acts as an integrator, ensuring alignment between technical specifications, design choices, and overall project goals. The Systems Engineer Agent operates within a layered architecture, integrating several key components: a knowledge source, a prompt engineering module, and an LLM execution engine. 

The agent's effectiveness is directly tied to the quality and accessibility of its knowledge base. This knowledge source represents the collective experience and expertise accumulated over years of practice. We represent this knowledge in the form of a knowledge graph, implemented as a CSV file containing nodes and relationships. This representation allows for structured storage and retrieval of information, facilitating informed decision-making by the agent. To manage the scale of the knowledge graph, the CSV content is split into manageable chunks, each associated with metadata such as chunk ID, source reference, and row range \cite{langchain_split_docs}. This chunking strategy optimizes the embedding and retrieval process.  Text chunks are then processed using an embedding model (OpenAI's text-embedding-3-small) to generate vector representations \cite{RAG}. These vectors are stored in a vector database for efficient similarity search.  This vector database approach enables rapid identification of relevant knowledge based on semantic similarity to the user's query \cite{senetence_BERT}. The use of vector embeddings allows the agent to understand the underlying meaning of the query and retrieve more contextually relevant information.  

At runtime, the Systems Engineer Agent converts user prompts into “search queries” through the prompt engineering module. This conversion often leverages the LLM itself to rephrase or expand upon the original prompt, ensuring that it effectively captures the user's intent.  The refined query is then used to search the vector database, retrieving the top-ranked and most relevant chunks. Thereafter, the retrieved chunks are appended to the user's original prompt, providing the agent with the necessary context to generate a comprehensive and informed response \cite{RAG}.  This process mimics the way a human systems engineer would consult internal documentation and prior experience to address a new design challenge. In Appendix \ref{sec:KG_information_retieval}, we empirically demonstrate examples of knowledge retrieval from the graph for verifying the RAG mechanism during the `Requirement Generation' step.

The Systems Engineer Agent performs several critical tasks within this framework. These include:
\begin{itemize}
    \item Technical Requirements Generation: Based on user prompts and leveraging its knowledge source, the agent generates technical requirements.  The contextual role is provided to the LLM agent to accomplish this task, ensuring consistency and adherence to established engineering principles. This capability streamlines the requirements elicitation process and minimizes the potential for ambiguity or omissions. One key advantage with using knowledge source for this task is the ability to formulate both functional and non-functional requirements simultaneously.
    \item Airfoil Design Review: The agent evaluates new airfoil designs, comparing them against established technical requirements and utilizing its accumulated design knowledge. The agent receives the airfoil profile generated by the Design Engineer along with the corresponding aerodynamic performance results.  Critically, the agent incorporates multi-modal vision models to assess airfoil profile plots and performance data visually. This capability allows for a more nuanced evaluation than relying solely on numerical data, accounting for factors such as visual symmetry, smoothness, and overall aesthetic appeal.  Feedback is provided to the Design Engineer, highlighting potential areas for improvement and suggesting alternative design approaches.
\end{itemize}
The agent's feedback on airfoil designs is driven by a dual consideration: alignment with technical requirements and leveraging its knowledge base for prior experience with design specifics. This knowledge base represents the accumulated understanding of design trade-offs, potential failure modes, and best practices. The agent uses this knowledge to anticipate potential issues, suggest proactive design improvements, and ensure that the final design meets the required performance characteristics. The agent uses Google's `gemini-2.5-flash' LLM model for its task accomplishment.

\subsection{Design Engineer Agent}
\label{sec:design_engg}
The Design Engineer agent represents a critical mid-level role within our multi-agent system designed to automate and enhance the airfoil design process. This agent is responsible for the iterative cycle of creating, analyzing, visualizing, and refining airfoil designs. The Design Engineer agent collaborates closely with the Systems Engineer agent to ensure design feasibility and adherence to overall system requirements to achieve performance targets. The core functionality of the Design Engineer agent centers around managing the design space, generating candidate airfoils, analyzing their aerodynamic performance, and iteratively refining designs based on feedback. This process is not a linear sequence but a continuous loop, with insights gained from analysis triggering adjustments to the design and further rounds of evaluation. Below we describe the core tasks and tools associated with this agent.

\subsubsection{Design Space Sampling}
The agent commences its task by sampling the design space, utilizing a suite of strategies to generate diverse candidate airfoils. The \emph{Sampling tool}, a custom-built Python function, provides options for Latin Hypercube, Sobol, Halton, and random sampling. Each strategy possesses unique characteristics suitable for different design exploration goals. The tool generates a set of parameter values for maximum camber, maximum camber location, and maximum thickness that define the 4-digit NACA airfoil geometry. 

\subsubsection{NACA Profile Generation}
Once the design parameters are sampled, the \emph{NACA profile generator tool} utilizes the standard 4-digit NACA airfoil equation to generate the corresponding airfoil coordinates. This equation, based on a mathematical representation of airfoil geometry, enables the creation of a wide variety of airfoil shapes by adjusting three key parameters: the location of the maximum camber, the maximum camber value, and the maximum thickness value. It is important to note that these initial designs are not intended to be final products but rather starting points for further refinement. 

\subsubsection{Airfoil Visualization}
The airfoil shapes generated are then visualized using the \emph{Plotting tool}, a Python function using airfoil coordinates to plot its shape. These visualizations are critical for review by the Systems Engineer agent, allowing for a qualitative assessment of each design against technical requirements, especially non-functional requirements.

\subsubsection{Aerodynamic Analysis with AeroSandbox and NeuralFoil}
Following visualization, the aerodynamic performance of the generated designs is evaluated using the \emph{AeroSandbox tool}, an open-source, Python-based framework \cite{aerosandbox_phd_thesis}. AeroSandbox uniquely integrates physics-based aerodynamic models, structural modeling, and multidisciplinary optimization within a unified computational environment. Unlike conventional CFD tools, AeroSandbox leverages low-fidelity aerodynamic models, such as vortex lattice methods and thin airfoil theory, coupled with automatic differentiation capabilities through tools like CasADi. This approach facilitates rapid design space exploration and gradient-based optimization of airfoils, wings, and complete aircraft configurations.

Complementing AeroSandbox is \emph{NeuralFoil}, a machine-learning-based airfoil analysis tool that utilizes a neural network trained on extensive XFOIL simulation data \cite{neuralfoil}. NeuralFoil’s predictive capabilities for aerodynamic coefficients ($C_l, C_d, C_m$) span a broad spectrum of Reynolds numbers and angles of attack, enabling near-instantaneous predictions with acceptable accuracy. By substituting numerical flow solutions with learned surrogates, NeuralFoil offers substantial computational savings, making it particularly well-suited for real-time applications, optimization loops, and scenarios where CFD simulations are impractical. This tool is primarily used for design analysis and optimization.

\subsubsection{Design Optimization}
Design optimization represents the final step in the design process where a candidate design chosen during the iterative design phase is optimized against the design objectives. The \emph{Optimization tool} utilizes Kulfan coordinates to represent and manipulate the airfoil geometry during the optimization stage \cite{kulfan}. Kulfan coordinates offer a distinct advantage over traditional Cartesian coordinates by providing a smoother and more stable representation of the airfoil shape, particularly when dealing with significant geometric changes. This property facilitates the optimization process by preventing abrupt shifts and ensuring a more controlled evolution of the design. The optimizer employs a multi-objective optimization routine, balancing aerodynamic performance metrics such as lift and drag coefficient, while adhering to other general constraints required for convergence.

\subsubsection{Design Selection and Revision}
The designs generated by the Design Engineer agent are reviewed by the Systems Engineer agent in an iterative process. The Design Engineer, upon receiving critical feedback about the designs then initiates a design selection and revision process, using his background knowledge and feedback received. This agent then modifies the design appropriately, and provides the new design sample for review again.
\\

The Design Engineer agent plays a critical role in iteratively refining designs based on Systems Engineer's feedback, technical requirements, and its own knowledge base. For example, if the Systems Engineer suggests a reduction in thickness value to aid with manufacturing, the Design Engineer might choose a lower maximum thickness value for generating the new airfoil shape guided by his knowledge. Such domain knowledge driven interaction between agents generates context-aware suggestions, considering not only the immediate performance goals but also the broader system requirements and manufacturing limitations. This integration allows the Design Engineer to operate with a higher level of intelligence and adaptability by continuously refining the design towards optimal performance and feasibility. The Design Engineer agent uses OpenAI's `gpt-4o' LLM model for its task accomplishment.

\subsection{Manager (Human User)}
\label{sec:manager}
Our framework incorporates a Manager role, representing the human user responsible for guiding the design exploration process. This individual does not directly manipulate the LLM-based design agents, but instead acts as a gatekeeper and evaluator, ensuring the generated designs align with overarching project goals, constraints, and evolving requirements. The Manager’s role is crucial for injecting domain expertise and addressing nuances that automated systems may overlook. The Manager’s involvement begins at the problem definition stage.  The initial design problem is formulated and communicated to the LLM-based design agents, setting the scope for subsequent exploration. During the final design selection phase, the Manager critically assesses a limited set of candidate designs, typically the top-ranked solutions chosen by the Design Engineer agent. This evaluation considers factors not easily quantifiable by the automated system, such as aesthetic preferences, manufacturability, and potential for future scalability. The Manager's authority extends to determining when the design cycle concludes. This termination condition is triggered by the Manager’s explicit acceptance of a design, indicating a satisfactory outcome has been achieved. Furthermore, the Manager’s ability to terminate the process allows for agile adaptation to changing project needs or the identification of fundamentally flawed design approaches. This provides a mechanism for course correction and ensures that the design process remains responsive to evolving requirements. Thus, this human-in-the-loop framework allows us to harness the power of LLM-based agents while retaining the crucial element of human expertise and judgment, facilitating a more robust and adaptable engineering design process. 

\subsection{Workflow}
\label{sec:workflow_desc}
In this section, we describe the sequence of operations in our multi-agent workflow shown in Figure \ref{fig:workflow}. The design workflow within this multi-agent system is structured around a main design sequence, orchestrated by the Manager, Systems Engineer, and Design Engineer agents, alongside an auxiliary sequence managed by a Graph Ontologist agent. The auxiliary knowledge graph (KG) generation sequence is executed prior to the main design sequence, establishing a foundational knowledge base for subsequent design iterations. This sequential approach aims to leverage the strengths of each agent with the Manager (human user) providing high-level direction, the Systems Engineer translating requirements, the Design Engineer generating and refining designs, and the Graph Ontologist providing structured knowledge. The main design sequence consists of six steps:

\begin{enumerate}
    \item Kickoff: The workflow initiates with a kickoff phase where the Manager provides an initial prompt to the Systems Engineer agent. This prompt outlines the problem goals and objectives, serving as the genesis of the design process. The initial prompt provided by the Manager is:
  \begin{textbox}{Manager prompt for Kickoff}
  \small
    “Develop a 4-series NACA airfoil that has the maximum lift to drag ratio at Mach=0.6. You can assume $Re=5\times10^6$ approximately and an AoA=0 deg for this phase of development.”
  \end{textbox}
  The selection of Mach=0.6, Re=$5\times10^6$, and AoA=0 represents typical test conditions for a range of applications, facilitating a relevant and focused design exploration .
  \item Requirement elicitation phase: Following the kickoff, the Systems Engineer agent undertakes requirement elicitation task. This crucial step transforms the initial prompt into a more detailed and technically grounded set of engineering requirements. The KG plays a pivotal role here, leveraging existing knowledge to elaborate on vague objectives and translate them into quantifiable design targets. For example, a general requirement like “design needs to minimize drag” might be transformed into “design must have a $C_d < 0.01$” through the KG's structured knowledge. Functional requirements, such as the aforementioned drag coefficient limit, can be directly assessed and evaluated. Non-functional requirements, like “the airfoil appearance must be aesthetically pleasing to the target customer,” are essential for comprehensive design consideration, though inherently more challenging to quantify and may involve qualitative assessments using our LLM agents. To illustrate this transformation, consider the initial prompt concerning maximum lift-to-drag ratio. The KG, enriched with aerodynamic principles and design best practices, can elicit specific functional requirements such as a target $C_l$ range or a maximum allowable $C_m$ value. Non-functional requirements might include constraints on manufacturing feasibility or aesthetic preferences.
  \item Design phase: The subsequent design phase is primarily managed by the Design Engineer agent. This involves the following sub-phases:
    \begin{enumerate}[label=(\alph*)]
        \item Design space sampling: A stage where the agent generates a set of design parameters: maximum camber, maximum camber location, and maximum thickness, based on a predefined strategy. We selected a sample size of 100 designs, justified by the need to explore a reasonable portion of the design space without overwhelming the subsequent analysis and review processes.
        \item Airfoil shape generation: The chosen design parameters are then fed into the NACA profile generator tool, producing airfoil profiles. The tool, by default, generates profiles with 200 points on both the top and bottom surfaces. 
        \item Airfoil visualization: Following generation, the plotting tool is employed to create visual representations of these profiles, facilitating intuitive understanding and communication throughout the design process.
        \item Airfoil analysis: The generated airfoil shapes are then subjected to aerodynamic analysis using NeuralFoil and AeroSandbox as discussed earlier. The primary Quantities of Interest (QoIs) assessed include lift coefficient ($C_l$), drag coefficient ($C_d$), and pitching moment coefficient ($C_m$). 
        \item Filtering: Following analysis, a filtering step is implemented. Airfoils with $C_l < 0.5$ are filtered out; this criterion reflects a practical engineering constraint related to the minimum lift requirement for stable flight and avoids designs that would be non-viable for intended applications.  The top 10 ranking designs are then progressed for design review, effectively reducing the burden on the Systems Engineer and aligning with time constraints commonly encountered in real-world engineering teams.
    \end{enumerate}
  \item Design review and feedback: The design review and feedback phase leverages the capabilities of `gemini-2.5-flash' multi-modal model. The multi-modal input capability is critical; it allows the Systems Engineer agent to analyze both the visual representation of the airfoil profiles and their corresponding performance data. For example, a mutli-modal LLM can be used to assess the smoothness of the airfoil’s leading edge or to correlate a sudden increase in $C_d$ with a specific visual anomaly in the profile. The Manager can also provide his feedback at this stage, aligning the design with broader project goals. This feedback is integrated into the design process, guiding the Design Engineer’s subsequent modifications. 
  \item Design selection and revision: Following the review, the Design Engineer selects the most promising design candidate, incorporating design change feedback from the Systems Engineer. This design then undergoes further review and modification in an iterative cycle until deemed valid by the Manager. The final design candidate is then passed to an optimization tool embedded within the NeuralFoil framework.
  \item Optimization:  The optimizer utilizes objectives and constraints derived from the technical requirements and pre-configured options to converge on an optimal solution. The optimization algorithm and specific objectives are crucial factors influencing the quality and efficiency of the final design.     
\end{enumerate}

\section{Experiments}
\label{sec:experiments}
This section details the key findings from our workflow, demonstrating the integrated functionality of LLM agents and human-in-the-loop interaction in an airfoil design process. We outline the sequence of events, from the initial generation of technical requirements to the final optimization of the airfoil shape. 
\vspace{5pt}

\textbf{Creation of Technical Requirements}\\
The design process commenced with Stage 2, the creation of technical requirements. Guided by a manager prompt, the Systems Engineer agent synthesized a comprehensive technical requirements document, as shown below. This document serves as a pivotal guiding principle for the entire team, outlining the functional and non-functional criteria the airfoil design needs to satisfy. The Systems Engineer agent's ability to draw upon its knowledge base, presumably informed by established engineering principles and design standards, ensures that the technical requirements were both relevant and complete. This initial step establishes a crucial foundation, minimizing ambiguity and setting clear objectives for the subsequent design phases. 

\begin{figure}[H]
    \centering
    \begin{textbox}{Functional Requirements}
    \scriptsize
    "- Objective Function:  The primary objective is to maximize the lift-to-drag ratio (Cl/Cd) of the airfoil section.

    - Airfoil Family: The airfoil design must strictly adhere to the NACA 4-digit series geometry.

    - Geometric parameters: 
    \begin{itemize}
        \vspace{-0.2cm} \item Maximum camber ($m$): The first digit of the NACA 4-digit series, representing the maximum camber as a percentage of the chord. The design space for `m` shall be constrained to $0\% <= m <= 9\%$. A non-zero camber is expected to generate lift at AoA=0.
        \vspace{-0.2cm} \item Location of maximum camber ($p$): The second digit of the NACA 4-digit series, representing the position of maximum camber from the leading edge in tenths of the chord. The design space for `p' shall be constrained to $0 <= p <= 9$ (0\% to 90\% chord).
        \vspace{-0.2cm} \item Maximum thickness ($t$): The last two digits of the NACA 4-digit series, representing the maximum thickness as a percentage of the chord. The design space for $t$ shall be constrained to $6\% <= t <= 20\%$ to manage drag at M=0.6 and avoid excessive drag coefficients.
    \end{itemize}
    - Operating conditions:
    \begin{itemize}
        \vspace{-0.2cm} \item Mach Number (M): The design must be optimized for a free-stream Mach number of 0.6.
        \vspace{-0.2cm} \item Reynolds Number (Re): The design must be optimized for a Reynolds number of approximately $5 \times 10^6$ (based on chord length)
        \vspace{-0.2cm} \item Angle of Attack (AoA): The design must be optimized for an angle of attack of 0 degrees.
    \end{itemize}
    - Performance metrics: The design must provide quantified values for the lift coefficient (Cl), drag coefficient (Cd), and the resulting lift-to-drag ratio (Cl/Cd) at the specified operating conditions."
  \end{textbox}
  \label{fig:functional_reqm}
\end{figure}

\vspace{-0.5cm}

\begin{figure}[H]
    \centering
    \begin{textbox}{Non-Functional Requirements}
    \scriptsize
    " - Manufacturability: The resulting airfoil geometry should be practical and feasible for manufacturing using conventional methods, avoiding features that would significantly increase production complexity or cost.

    - Aerodynamic performance
    \begin{itemize}
        \vspace{-0.2cm} \item Critical Mach Number: The designed airfoil should aim to have a critical Mach number greater than 0.6 to minimize the onset of significant compressibility effects (e.g., wave drag) at the operating Mach number. The designed airfoil should aim to have a critical Mach number greater than 0.6 to minimize the onset of significant compressibility effects (e.g., wave drag) at the operating Mach number.
        \vspace{-0.2cm} \item Boundary Layer Characteristics: The design must consider the boundary layer transition and separation characteristics at $Re=5\times10^6$ to minimize profile drag and ensure stable flow.
        \vspace{-0.2cm} \item Pressure Distribution: The airfoil geometry should result in a favorable pressure distribution along the chord that contributes to high lift and low drag at AoA=0 and M=0.6.
    \end{itemize}
    
    - Analysis Compatibility: The airfoil geometry must be compatible with standard computational fluid dynamics (CFD) or panel method tools (e.g., XFOIL) for accurate aerodynamic performance evaluation.
    
    - Robustness: The design should ideally exhibit a reasonable performance envelope, meaning that minor deviations from the specified operating conditions (M=0.6, Re=5e6, AoA=0) do not lead to drastic performance degradation."
  \end{textbox}
  \label{fig:non_functional_reqm}
\end{figure}
\vspace{5pt}

\begin{figure}[h]
    \centering
    \includegraphics[width=1\linewidth]{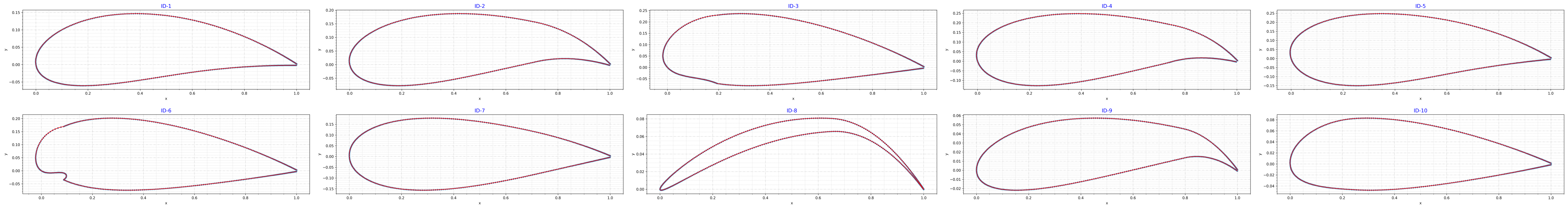}
    \caption{Initial airfoil samples generated by the Design Engineer agent using the NACA generator and plotter tool. The image shows 10 out of the 100  samples generated in this step following parameter selection in Stage 2.}
    \label{fig:airfoil_vis}
\end{figure}

\textbf{Initial Design Generation and Visualization}\\
Stage 3 then initiated the design generation phase. The Design agent employed Latin Hypercube sampling to generate an initial set of 100 diverse airfoil designs. The Design agent then proceeded to generate and visualize the airfoil shapes themselves. This visualization step, as depicted in Figure \ref{fig:airfoil_vis}, allows for an intuitive understanding of the design variations and facilitates visual assessment by Systems Engineer agent and Manager during the design review phase. 

\begin{figure}[h]
    \centering
    \includegraphics[width=1\linewidth]{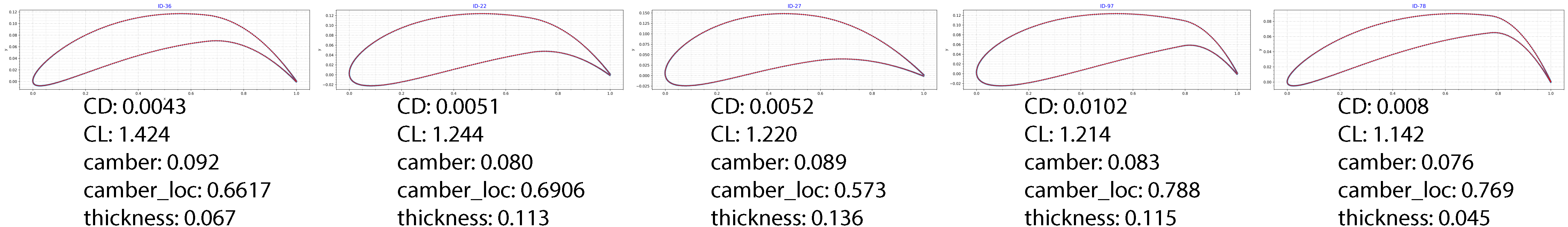}
    \caption{Initial filtered design samples selected for design review by the Systems Engineer agent.}
    \label{fig:filtered_airfoils}
\end{figure}

\textbf{Design analysis and filtering}\\
Following the generation of initial designs, a design analysis and filtering process was implemented to reduce the number of designs for the initial review. Each design was evaluated for its aerodynamic performance using NeuralFoil. Following this, the Design agent excluded designs exhibiting a coefficient of lift ($C_l$) below 0.5, a guideline established to ensure basic aerodynamic performance. This filtering step is accomplished by a python function that allows designs to be sorted and filtered. Figure \ref{fig:filtered_airfoils} shows the airfoil samples selected for design review. We note that airfoils with low thickness values exhibit higher lifts and consequently appear at the top of this filtered list. However, due to difficulties in meeting non-functional requirements such as manufacturability, some of these designs are subsequently considered invalid during the review process (by the Systems Engineer).

\textbf{Initial Design Review}\\
In stage 4, the Systems Engineer agent leverages its expertise derived from a larger knowledge base along with a vision LLM to evaluate the proposed designs against the previously defined technical requirements. This evaluation involves a nuanced consideration of aerodynamic performance and adherence to the technical specifications, especially non-functional requirements that are qualitative. The Systems Engineer provides clear feedback on each design, categorizing them as either “invalid” (unable to meet technical requirements) or ``valid''. This structured feedback mechanism is crucial for guiding the Design Engineer agent’s subsequent revisions and ensuring that the design process converges on a good design candidate. Below we show feedback generated by the Systems Engineer agent for initial design samples from Stage 4. Additionally, in Appendix \ref{sec:agent_reasoning}, we provide an example of the background reasoning and task breakdown used by the Systems Engineer agent during the design review process of a sample airfoil profile and aerodynamic data.

\begin{textbox}{Feedback on Design ID-22}
  \scriptsize
    "The airfoil profile (ID-22) visually presents a smooth, cambered shape with conventional leading and trailing edges, indicative of a well-formed NACA 4-digit series airfoil. The aerodynamic data shows exceptional performance with a lift coefficient (Cl) of 1.244 and a very low drag coefficient (Cd) of 0.00508, resulting in an outstanding lift-to-drag ratio (Cl/Cd) of 244.88 at the specified operating conditions $(M=0.6, Re=5\times10^6, AoA=0)$. The geometric parameters camber (0.08015), camber location (0.69065), thickness (0.11287) are all within the defined design space and contribute effectively to this high performance. To potentially improve the design further, consider exploring a slightly more forward camber location (e.g., 40-50\% chord) as moving camber towards the leading edge can increase lift, which might further enhance the Cl/Cd ratio. Additionally, a detailed pressure distribution analysis and critical Mach number assessment would provide deeper insights for optimization.","Valid":true
    \vspace{0pt}
    \begin{center}
        \includegraphics[width=0.5\textwidth]{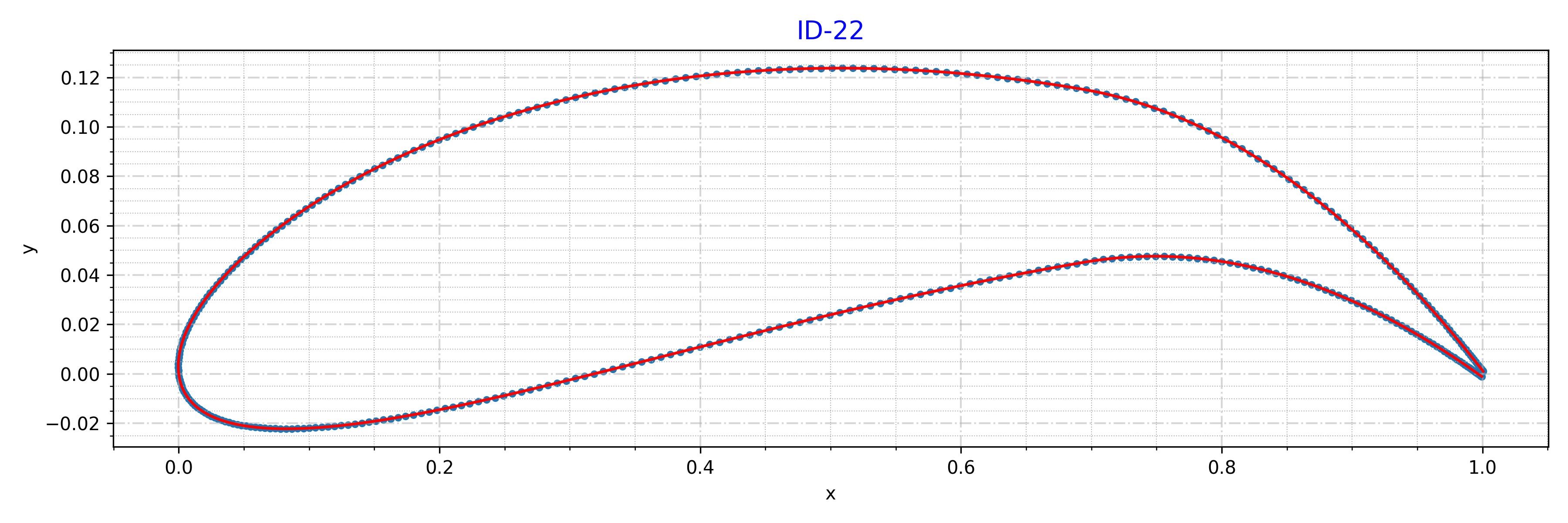}
    \end{center}
  \end{textbox}

\vspace{2pt}
\begin{textbox}{Feedback on Design ID-36}
  \scriptsize
    "The airfoil profile, as depicted in the image, shows a well-defined shape with a noticeable camber, consistent with a lifting airfoil. The current design exhibits excellent aerodynamic performance with a very high lift coefficient $(Cl = 1.42)$ and an exceptionally low drag coefficient $(Cd = 0.0043)$, resulting in an outstanding lift-to-drag ratio of 330.79 at the specified operating conditions $(M=0.6, Re=5\times10^6, AoA=0)$. The geometric parameters are: camber: 9.2\%, camber location: 66.2\% chord, and thickness: 6.7\% chord. While the thickness and camber location are within the specified design space, the maximum camber of 9.2\% slightly exceeds the functional requirement's upper limit of 9\%. To strictly adhere to the NACA 4-digit series geometry constraint, the maximum camber should be adjusted to be at most 9\%. Further optimization could explore moving the camber location slightly forward (e.g., 40-50\% chord) to potentially enhance lift characteristics or refine pressure distribution, and a critical Mach number analysis is recommended to confirm compliance with the non-functional requirement.","Valid":false
    \vspace{0pt}
    \begin{center}
        \includegraphics[width=0.5\textwidth]{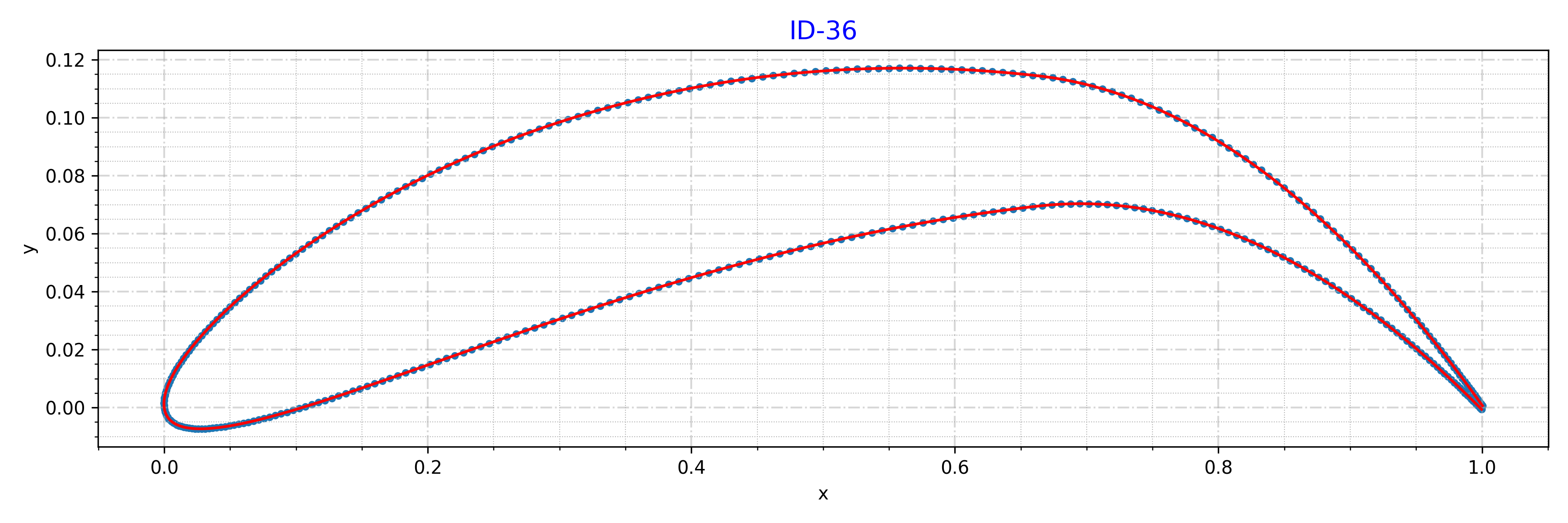}
    \end{center}
  \end{textbox}

\vspace{2pt}
\begin{textbox}{Feedback on Design ID-78}
  \scriptsize
    "The airfoil profile ID-78, visually appearing very thin and cambered, demonstrates exceptional aerodynamic performance with a high lift coefficient (Cl=1.14) and a very low drag coefficient (Cd=0.0085), resulting in an outstanding Cl/Cd ratio of 134.45. This performance is largely attributed to the significant camber (7.62\%) at an aft location (76.97\% chord). However, the design is critically flawed as its maximum thickness (4.59\%) falls below the minimum functional requirement of 6\% (design space for 't' is $6\% <= t <= 20\%$). This extreme thinness, while potentially contributing to a high critical Mach number, raises significant concerns regarding manufacturability and structural integrity. To improve the design, the maximum thickness should be increased to at least 6\%, ideally targeting 6-8\%, to comply with the requirements while maintaining good aerodynamic efficiency.","Valid":false
    \vspace{0pt}
    \begin{center}
        \includegraphics[width=0.5\textwidth]{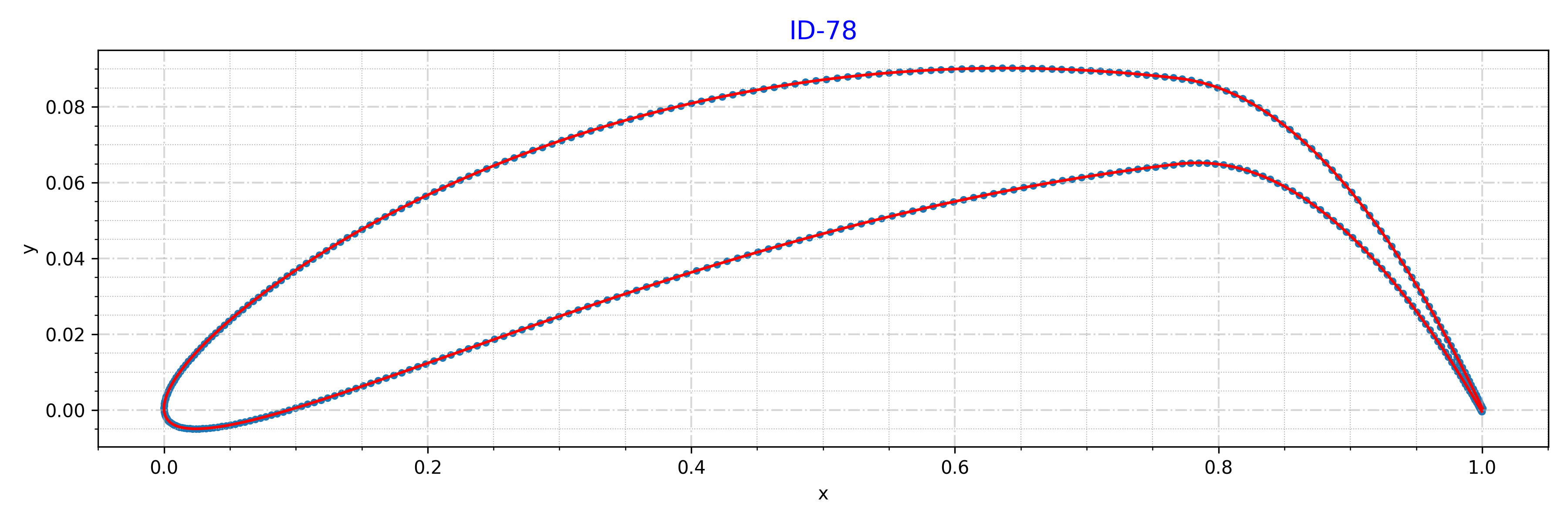}
    \end{center}
  \end{textbox}
  
\textbf{Iterative Design Revision and Review}\\
Based on the Systems Engineer's feedback and its design-related knowledge base, the Design Engineer agent selects one design for further refinement. The Design Engineer then modifies the design parameters according to the Systems Engineer’s suggestions, visualizes the resulting airfoil, and evaluates its performance using the AeroSandBox tool. This chosen design is then presented to the Systems Engineer agent along with Manager for an iterative review-revision process. The response from the agent below shows the extracted knowledge from the design KG, reasoning used behind the choice of best candidate design and the revisions made to the design.
\begin{textbox}{Design revision 1: Designer's reasoning}
  \scriptsize
    **Design Parameters:**
        - Max Camber: 9\% (0.09)
        - Camber Location: 50\% (0.5)
        - Thickness: 12\% (0.12)\\
        
**Reasoning:**
- The maximum camber is set to 9\% to adhere to the functional requirement's upper limit while maximizing lift.
- The camber location is moved to 50\% chord to potentially enhance lift characteristics, as suggested by feedback and supported by the knowledge base.
- The thickness is set to 12\% to ensure manufacturability and maintain structural integrity, while also being within the specified design space.\\

 \textbf{**Knowledge Source Reference:**}
 \emph{
- Moving the camber location towards the leading edge increases lift.
- Higher camber values can increase the lift coefficient but may also increase drag due to flow separation.
- Thickness affects drag and manufacturability.}

The revised airfoil design has been generated using the `generate airfoil' tool with the specified parameters.
    \vspace{0pt}
    \begin{center}
        \includegraphics[width=0.5\textwidth]{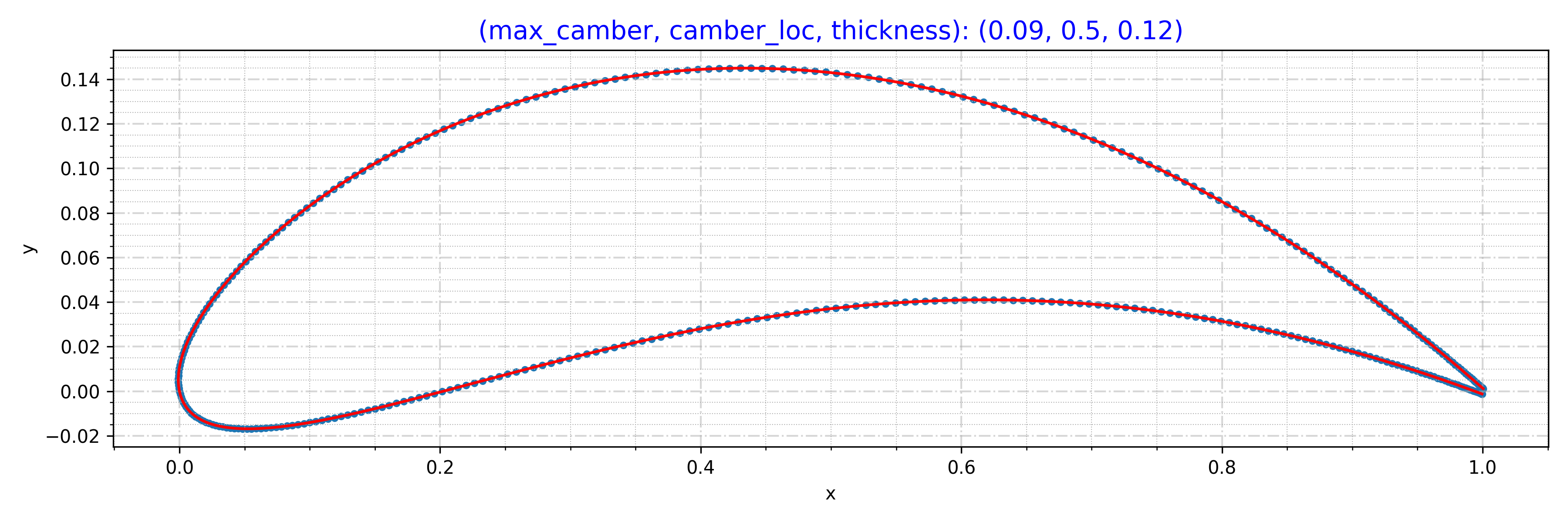}
    \end{center}
  \end{textbox}

This is followed by a design review jointly by the Systems Engineer and the Manager. The feedback provided by the Systems Engineer is shown below, where the design specification provided by the Designer is compared against the Functional requirement of meeting the NACA profile standards. Additionally, the Systems Engineer also identifies areas of improvement such as adjustments to camber location to help improve the pitching moment. 
\begin{textbox}{Design revision 1: Systems Engineer feedback}
  \scriptsize
    "The airfoil profile, identified as NACA 9512 from the geometric parameters (camber: 0.09, camber location: 0.5, thickness: 0.12), exhibits a smooth, cambered shape typical of the NACA 4-digit series. The aerodynamic performance is exceptional, with a high lift coefficient (Cl = 1.135) at AoA=0 and a very low drag coefficient (Cd = 0.0055), yielding an outstanding lift-to-drag ratio of 206.24. All geometric parameters adhere to the specified design space constraints. For further improvement, it is critical to verify the critical Mach number is indeed greater than 0.6 to avoid compressibility effects at the operating Mach number. Additionally, optimizing the camber location could help manage the pitching moment (CM = -0.272) and a detailed boundary layer analysis could further refine drag reduction.", "Valid": true
  \end{textbox}
  
At this point, the Manager intervenes in the review process and provides his/her feedback about the proposed design. This human-in-loop arrangement allows a human to supervise the design process, while the autonomous agents perform a major chunk of work.
\begin{textbox}{Design revision 1: Manager feedback}
  \scriptsize
    "I want the design to have standard camber of between 1-6\% while maximizing L/D ratio and optimizing CM. Can you update your response accordingly and mark this design as invalid"
  \end{textbox}

Based on this human feedback, the Systems Engineer updates its feedback for the Design Engineer with the updated requirement, while flagging the design as invalid to continue design iteration.

\begin{textbox}{Design revision 1: Systems Engineer updated feedback}
  \scriptsize
    "Thought: \emph{The user has provided new feedback for the design. I need to incorporate this feedback into my analysis and update the `Valid` status to `False` for the current design (`revision1'). The key changes are:}
    
    \emph{**Camber constraint**: The design should have a standard camber between 1-6\%. The current design has a camber of 9\%. This violates the new constraint.}\\
    
    The airfoil profile, identified as NACA 9512 from the geometric parameters (camber: 0.09, camber location: 0.5, thickness: 0.12), exhibits a smooth, cambered shape typical of the NACA 4-digit series. However, this design is **invalid** as the maximum camber of 9\% violates the updated user requirement for camber to be between 1\% and 6\%. While the aerodynamic performance (Cl = 1.135, Cd = 0.0055, Cl/Cd = 206.24) is excellent, the design must adhere to the new camber constraint. For future iterations, the design should target a camber within the 1-6\% range. Additionally, while maximizing the L/D ratio, the pitching moment (CM = -0.272) should also be optimized, potentially by adjusting the camber location or thickness distribution to achieve a more favorable CM.", "Valid": false
  \end{textbox}

With this feedback, the Design Engineer agent revises the design to include improvements suggested. The output below shows the reasoning process followed.
\begin{textbox}{Design revision 2: Designer's reasoning}
  \scriptsize
    "**Design Parameters:**
- Maximum Camber: 0.05 (5\%)
- Camber Location: 0.4 (40\% chord)
- Thickness: 0.12 (12\%) \\

**Reasoning for Design Modification:**
- The maximum camber was reduced to 5\% to comply with the updated requirement of camber being between 1\% and 6\%.
- The camber location was set at 40\% of the chord to potentially improve lift and overall airfoil performance, as moving the camber location towards the leading edge increases lift.
- The thickness was maintained at 12\% to manage drag effectively at Mach 0.6 while avoiding excessive drag coefficients. \\

\textbf{**Knowledge Source Reference:**}
- Moving the camber location towards the leading edge increases lift.
- A more aft maximum camber location improves airfoil performance.
- Thickness should be within 6\% to 20\% to manage drag at Mach 0.6.

The revised airfoil design, identified as Revision 2, has been generated and saved. This design adheres to the updated constraints and aims to maximize the lift-to-drag ratio while optimizing the pitching moment."
    \vspace{0pt}
    \begin{center}
        \includegraphics[width=0.5\textwidth]{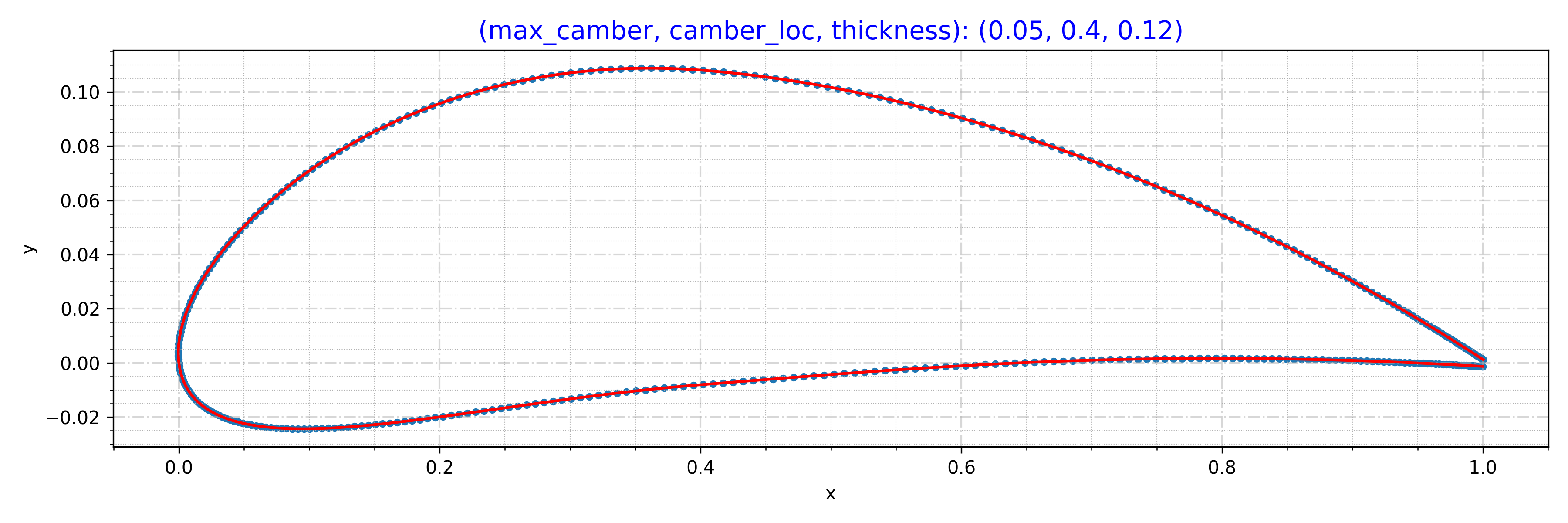}
    \end{center}
  \end{textbox}

The Systems Engineer feedback on Design Revision 2 is as follows:
\begin{figure}[H]
    \centering
    \begin{textbox}{Design revision 2: Systems Engineer feedback}
    \scriptsize
    "The airfoil profile, identified as a NACA 5412 based on the geometric parameters (camber: 0.05, camber location: 0.4, thickness: 0.12), exhibits a well-defined shape with a rounded leading edge and sharp trailing edge, typical for a lifting airfoil. The aerodynamic performance is excellent, with a high lift coefficient (Cl: 0.582) at 0 degrees angle of attack and a remarkably low drag coefficient (Cd: 0.00578), resulting in an outstanding lift-to-drag ratio of 100.71. This design adheres to all specified functional requirements, including the NACA 4-digit family, geometric parameter constraints, and operating conditions. The Cl value is above 0.5, making the design valid by the rule of thumb. For further improvement, it is crucial to verify the critical Mach number to ensure it is greater than 0.6 at the operating Mach number of 0.6 to mitigate compressibility effects. Additionally, a detailed pressure distribution analysis and robustness study at slightly off-design conditions would enhance the design's confidence." ``Valid": True
  \end{textbox}
\end{figure}

The Manager intervenes again here to suggest design modification in order to improve the lift coefficient further.
\begin{textbox}{Design revision 2: Manager feedback}
  \scriptsize
    "Can you suggest improvement to make Cl close to 0.7? Please mark this design as invalid"
  \end{textbox}

The Systems Engineer agent then updates its feedback for the Design Engineer to include the Manager's comments for improving Cl values further
\begin{figure}[H]
    \centering
    \begin{textbox}{Design revision 2: Systems Engineer updated feedback}
    \scriptsize
    "The current airfoil (NACA 5412) provides a lift coefficient (Cl) of 0.582, which is good, \emph{but the user has requested an improvement to achieve a Cl closer to 0.7}. To increase the lift coefficient at 0 degrees angle of attack, it is recommended to increase the maximum camber of the airfoil. For instance, exploring NACA 6412 (camber: 0.06, camber location: 0.4, thickness: 0.12) or NACA 7412 (camber: 0.07, camber location: 0.4, thickness: 0.12) could help achieve the target Cl while remaining within the NACA 4-digit series constraints. This adjustment would likely increase Cl, but further analysis would be needed to ensure the lift-to-drag ratio remains optimized and the critical Mach number requirement is still met. Valid: False"
  \end{textbox}
\end{figure}

The Design Engineer acts on this feedback to update the design further to meet the new request from the Manager
\begin{figure}[H]
    \centering
    \begin{textbox}{Design revision 3: Designer's reasoning}
    \scriptsize
    "To address the feedback and requirements, I need to determine the best design parameters for a revised airfoil that meets the specified criteria. The goal is to achieve a lift coefficient (Cl) closer to 0.7 while maximizing the lift-to-drag ratio (Cl/Cd) at the specified operating conditions. The feedback suggests exploring NACA 6412 and NACA 7412, which have higher camber values than the current NACA 5412. \\

    The requirements specify:
    - **Maximum Camber (m)**: 0\% to 9\%
    - **Position of Maximum Camber (p)**: 0\% to 90\%
    - **Maximum Thickness (t)**: 6\% to 20\% \\
    
    Given the feedback and requirements, I will consider the following:
    - Increasing the maximum camber to improve the lift coefficient.
    - Keeping the camber location at 0.4 to maintain a balance between lift and drag.
    - Maintaining the thickness at 0.12 to avoid excessive drag. \\
    
    I will generate a revised airfoil design using the `generate airfoil' tool with the following parameters:
    - Maximum Camber: 0.07 (for NACA 7412)
    - Camber Location: 0.4
    - Thickness: 0.12"
  \end{textbox}
\end{figure}

\begin{figure}[h]
    \centering
    \includegraphics[width=0.75\linewidth]{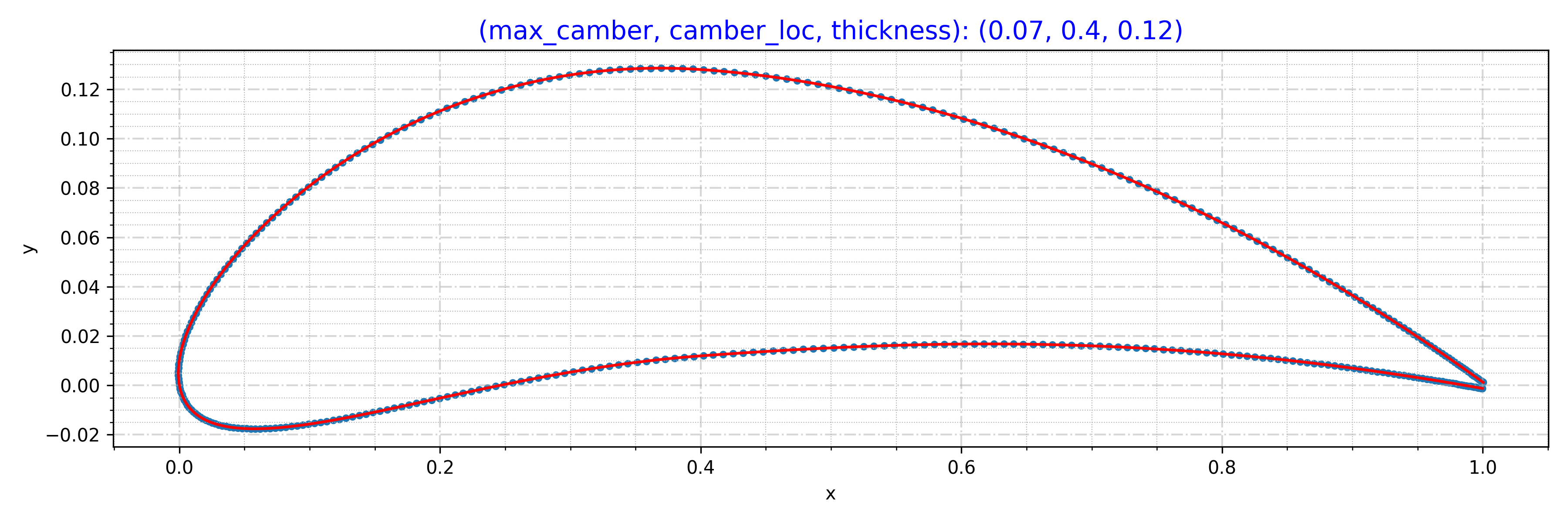}
    \caption{Final candidate design chosen for the optimization process after the iterative design phase. Note that the camber value for this design exceeds the 6\% criteria defined by the Manager at design iteration 1. However, due to the new request from the Manager to improve $C_l$ values further, the agents identify the need to increase the camber values beyond this range to meet the competing requirements of higher lift and pitching moment.}
    \label{fig:optimization_candidate}
\end{figure}

The iterative review and design process, spanning Stages 4 and 5, continues until a satisfactory design candidate is selected by the Manager. This loop, involving the Systems Engineer and Manager, ensures that the design not only met the technical requirements but also aligned with broader human user's goals and considerations. The final design candidate proposed by the Design Engineer is shown in Figure \ref{fig:optimization_candidate}.

\begin{figure}[h]
    \centering
    \includegraphics[width=0.75\linewidth]{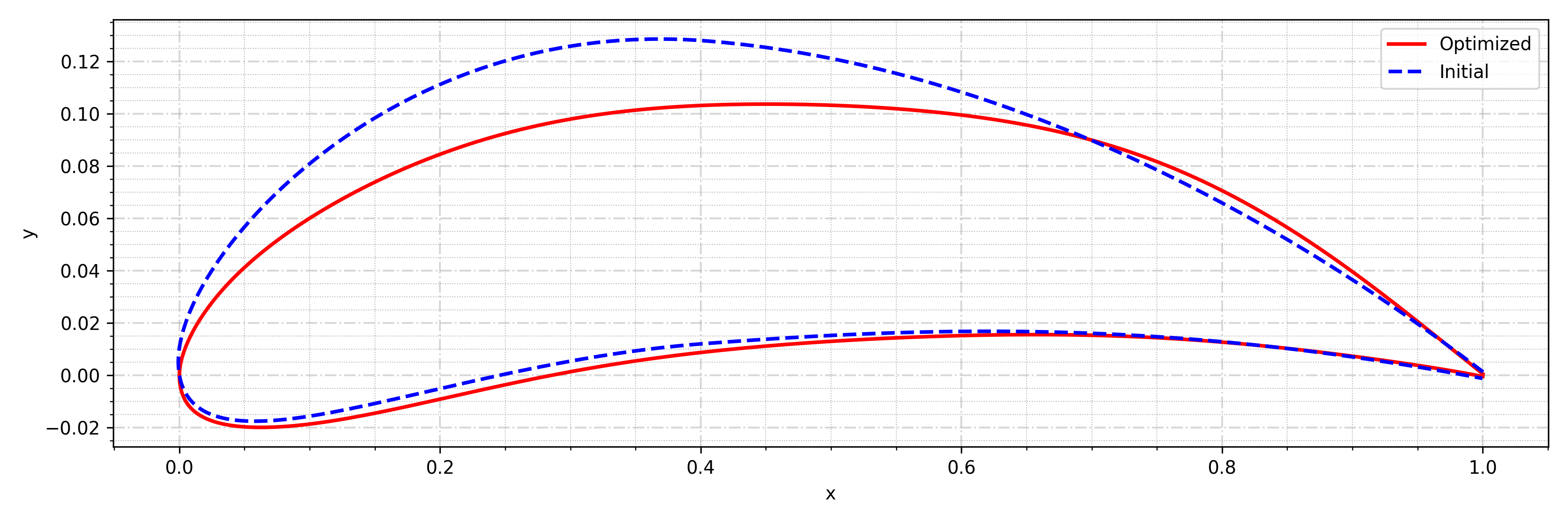}
    \caption{Comparison between optimized design and initial starting design selected through the iterative design process. The optimized design has performance values $C_l = 0.811$, and $C_d = 0.0044$ against the initial starting design with $C_l = 0.8077$ and $C_d = 0.0061$. Note that since the optimization process uses an initial design that satisfies (to a large extent) both functional and non-functional requirements, the end design does not suffer from qualitative shortcomings such as thin trailing edge which may lead to other challenges in manufacturing and robustness.}
    \label{fig:optim_vs_initial}
\end{figure}

\textbf{Design Optimization}\\
Stage 6 marks the final phase: design optimization. The design selected from the iterative review process serves as the starting point for this optimization effort. The optimization objective is defined as maximizing the lift-to-drag ratio (L/D) at a Mach number (Ma) of 0.6, Reynolds number (Re) of $5\times10^6$, and angle of attack (AoA) of 0 degrees. This objective function directly reflects a critical performance metric for airfoil designs – maximizing lift while minimizing drag. The design variables are the coordinates of the airfoil profile. For this phase, the Design Engineer utilizes the Optimization tool available inside NeuralFoil and AeroSandbox, where two default constraints for minimum local thickness and trailing edge values are enforced to ensure convergence to a feasible solution. Figure \ref{fig:optim_vs_initial} shows a comparison between the initial design and the optimized design obtained at the end of this stage.

Once an optimized design is obtained, the design process concludes at this point. Further iterations may be necessary to refine the design for final product realization, potentially involving structural and CFD analyses, which is considered beyond the scope of this present study. 

\section{Limitations and Future Works}
\label{sec:limitations}
The presented framework, while demonstrating the potential for integrating multi-agent systems (MAS) into iterative engineering design processes, contains certain limitations. Our approach currently embodies an academic exploration of this integration, focusing on a simplified representation of engineering tasks. Engineering design, by its inherent nature, frequently involves intricate and nuanced decision-making processes that extend beyond the scope of our current implementation. While the agents exhibited adaptability in navigating decision-making scenarios with the help of the knowledge graph, maintaining consistency across sophisticated engineering tasks remains a significant challenge. The problem formulation used in our work shows a relatively straightforward pathway, which is often a simplification compared to the complexities encountered in real-world engineering design. Unexpected sub-processes and iterations are common occurrences during the design cycle, deviating from the linear progression assumed in our demonstration. 

Furthermore, the tools leveraged by the Design Engineer agent are presently implemented as simple, Python-based modules, facilitating ease of integration within the workflow. However, a major impediment to broader adoption lies in the reliance on proprietary software crucial to engineering practice. Seamless integration with industry-standard Computer-Aided Design (CAD) and Computer-Aided Engineering (CAE) tools presents a considerable hurdle for our MAS framework. The complexity and closed nature of these tools often preclude direct interaction with agents, necessitating the development of intermediary layers and custom interfaces, a task that demands substantial resources and expertise. Future work will focus on addressing these limitations, exploring more complex workflows that incorporate agents to accomplish Computational Fluid Dynamics (CFD) analyses on final design.

In the future, a key area of development will be the agents’ ability to dynamically create and manage tools within the design workflow. Currently, our agents operate within a pre-defined toolset. The ability to generate tools on demand – for example, custom scripts for data analysis or specialized modules for simulation – would significantly enhance the framework’s flexibility and utility. This dynamic tool creation capability represents a substantial challenge, requiring advancements in automated code generation and agent reasoning, but is important to enable more autonomous and adaptive engineering design processes. We believe this capability will be pivotal in realizing the full potential of MAS within the engineering domain.

\section{Summary}   
\label{sec:summary}
In this work, we propose a novel multi-agent framework that leverages large language models (LLMs) to facilitate and accelerate the solution for a pedagogical challenge of designing high-performance NACA airfoils. We focus on the design of 4-digit NACA airfoils, for demonstration purposes. This framework comprises three specialized AI agents, each playing a distinct role in the iterative design cycle. The core agents are: a Graph Ontologist, a Design Engineer, and a Systems Engineer. The Graph Ontologist utilizes a prompt engineering approach to curate specialized knowledge graphs by extracting relevant information related to airfoil design and development from scientific literature. The extracted data is structured and organized within the graph, enabling efficient retrieval and utilization by other agents. Both Design and Systems Engineer are provided with their own customized knowledge graph that is suited for their roles and expertise. The Design Engineer is responsible for generating and modifying airfoil designs based on the information contained within the knowledge graph in consultation with the Systems Engineer agent. The Systems Engineer reviews designs generated by the Design Engineer, provides feedback for improvement, and ensures alignment with technical requirements. This agent also receives input from a human Manager to translate user decisions into actionable design feedback. 

The Design and Systems Engineer agents accomplish the design process through an iterative process. The Design Engineer, guided by its knowledge graph and design tools, produces an initial set of airfoil designs. The Systems Engineer then assesses these designs, evaluating both their quantitative performance characteristics (e.g., lift and drag coefficients) and qualitative attributes (e.g., manufacturability, robustness) as defined in the technical specification. This feedback is translated into actionable design improvement strategies, such as adjustments to airfoil geometry. For example, if the Systems Engineer identifies a potential issue with the airfoil’s stall characteristics, they might instruct the Design Engineer to increase the airfoil’s thickness or modify its leading-edge curvature.  The agents access and utilize their individual knowledge graphs, ensuring consistency and facilitating knowledge transfer between design iterations. To ensure that the iterative process converges on a suitable design, a human Manager is included in the final design selection phase. The Manager determines whether a design is valid based on his requirements. This process is repeated until a design that meets the Manager's requirements is obtained.

This synergistic approach prioritizes both quantitative and qualitative requirements during design iteration, ensuring that the final design meets overall acceptance criteria. In future work, we  will investigate the use of additional agents such as a CFD analyst to help address next steps in the design sequence along with methods to generate simple design tools dynamically during the process. In summary, the framework presented here demonstrates a path toward automating and augmenting the engineering design process, potentially improving design quality and accelerating innovation. 

\section*{Acknowledgments}
\vspace{-8pt}
This research was conducted using computational resources and services at the Center for Computation and Visualization, Brown University.

\section*{Funding}
\vspace{-8pt}
VK and GEK acknowledge support from Defense Advanced Research Projects Agency (DARPA) under the Automated Prediction Aided by Quantized Simulators (APAQuS) program, Grant No. HR00112490526, AFOSR Multidisciplinary Research Program of the University Research Initiative (MURI) grant FA9550-20-1-0358, ONR Vannevar Bush Faculty Fellowship (N00014-22-1-2795), and U.S. Department of Energy project SEA-CROGS (DE-SC0023191).

% \section*{Data and code availability}
% \noindent All codes will be made publicly available at {\small\url{https://github.com/varunsingh88/PET_bottle_buckling.git}} upon publication.

\bibliographystyle{elsarticle-num} 
\begin{footnotesize}
\bibliography{references}
\end{footnotesize}

\clearpage
\newpage
\makeatletter
\renewcommand \thesection{A\@arabic\c@section}
\renewcommand\thetable{A\@arabic\c@table}
\renewcommand \thefigure{A\@arabic\c@figure}
\makeatother

\section*{\Large{Appendices}}

\setcounter{figure}{0}
\setcounter{table}{0}
\setcounter{section}{0}
\setcounter{page}{1}

\section{Knowledge Graph information retrieval}
\label{sec:KG_information_retieval}
To validate and demonstrate the agent's ability to effectively retrieve and apply knowledge, we implemented a mechanism to trace the provenance of its generated outputs. The System Engineer agent's instructions were augmented to require explicit citation for any information retrieved from the knowledge graph via the Retrieval-Augmented Generation (RAG) framework. This modification compels the agent to label knowledge within its response using the format `\emph{(src:KG, retrieved knowledge info)}', providing a transparent link between its reasoning and the underlying knowledge base. We illustrate this capability through two examples from the technical requirements generation task.

\begin{figure}[h]
    \centering
    \includegraphics[width=1\linewidth]{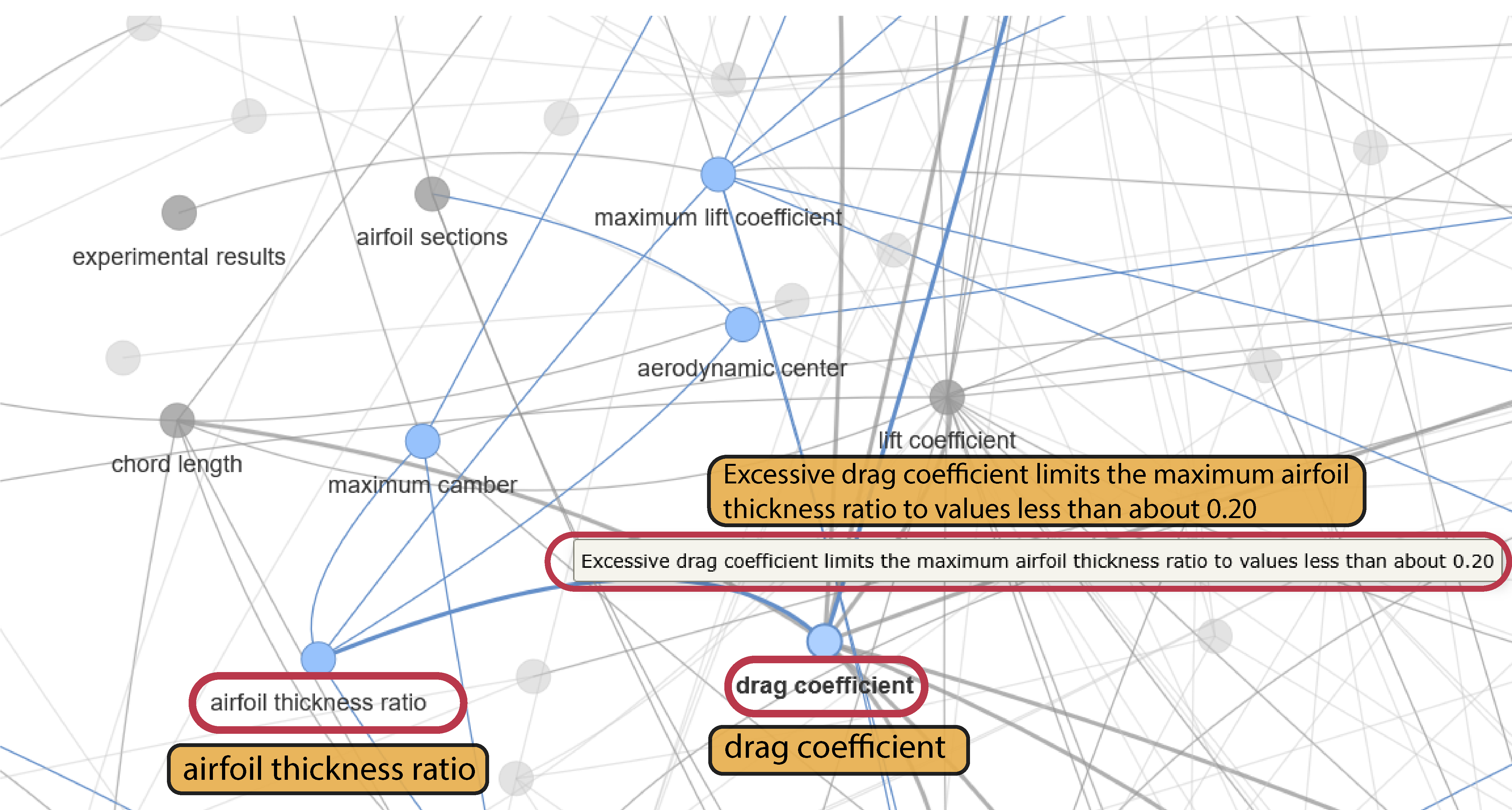}
    \caption{Section of the Systems Engineer's Knowledge Graph showing the two relevant nodes `drag coefficient' and `airfoil thickness ratio' retrieved to extract the required information using the ontological relationship between the two nodes for defining the maximum permissible airfoil thickness. The LLM prompt is used to extract semantically similar chunks from the graph data which is used as additional context for the final LLM response to the task.}
    \label{fig:KG_retieval_thickness}
\end{figure}

First, we examine the derivation of a quantitative constraint for airfoil thickness. Responding to a general directive to use NACA 4-series designs, the agent retrieved a critical relationship from the knowledge graph connecting the nodes `airfoil thickness ratio' and `drag coefficient'. The specific fact, "Excessive drag coefficient limits the maximum airfoil thickness ratio to values less than about 0.20," was directly utilized to establish an upper bound for the design parameter. As shown in the agent's response, this led to the concrete requirement constraining thickness `t' to $\leq 20\%$. This process, visually corroborated by the graph excerpt in Figure \ref{fig:KG_retieval_thickness}, showcases the agent's capacity to ground design parameters in fundamental principles stored within the KG.
\begin{figure}[h]
    \centering
    \begin{textbox}{Knowledge Graph trace during requirement generation for thickness}
    \scriptsize
    "**Maximum Thickness (t)**: The last two digits of the NACA 4-digit series, representing the maximum thickness as a percentage of the chord. The design space for `t` shall be constrained to $6\% \leq t \leq 20\%$ to manage drag at M=0.6 and avoid excessive drag coefficients (\emph{src:KG, airfoil thickness ratio,  drag coefficient: Excessive drag coefficient limits the maximum airfoil thickness ratio to values less than about 0.20}. Reasoning: This provides a practical upper limit for the thickness ratio to avoid excessive drag, which is crucial for maximizing L/D)."
  \end{textbox}
  \label{fig:functional_reqm}
\end{figure}

\begin{figure}[H]
    \centering
    \includegraphics[width=1\linewidth]{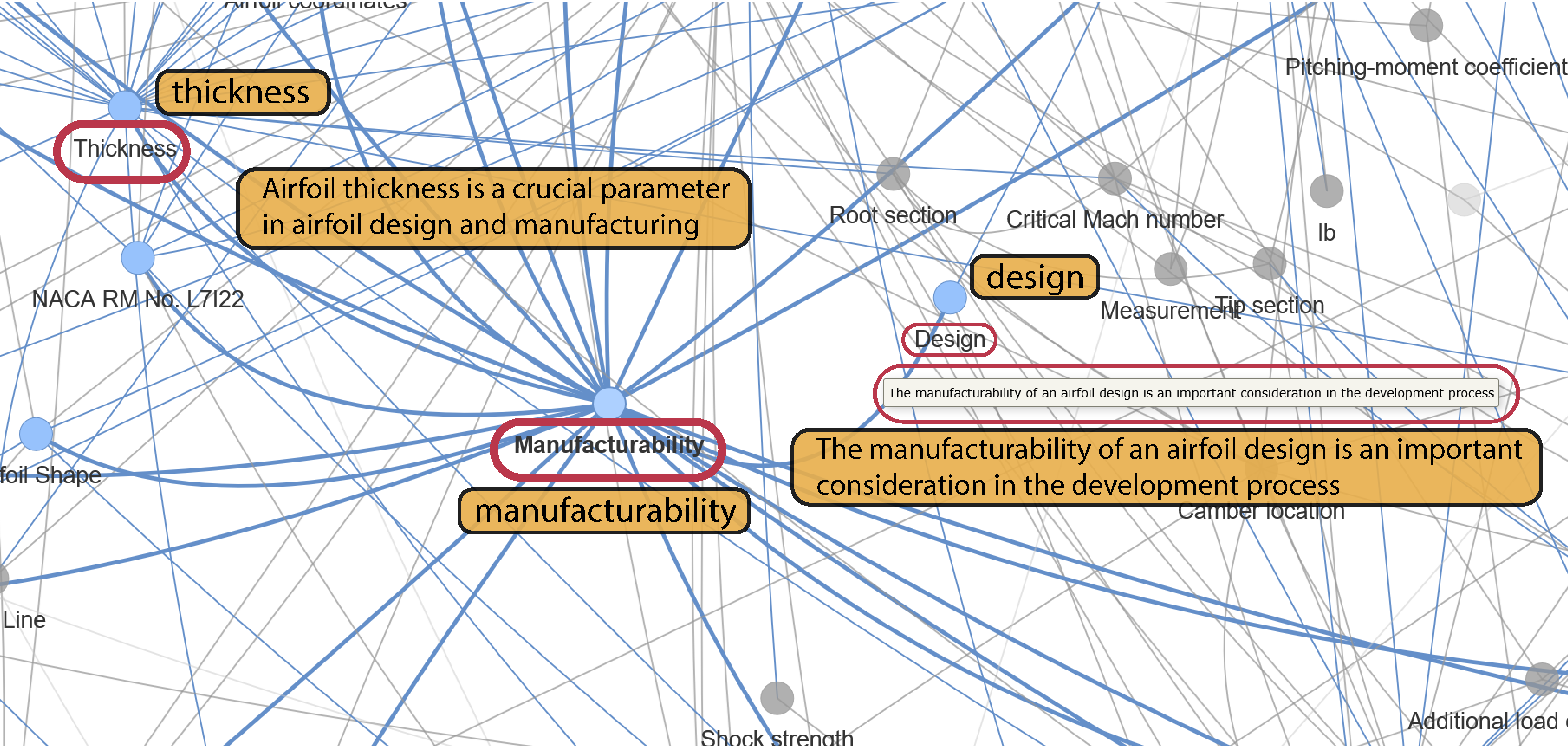}
    \caption{Section of the Systems Engineer's Knowledge Graph showing the two relevant nodes `Design' and 'Manufacturability' retrieved. The LLM prompt is used to extract semantically similar chunks from the graph data, from which the LLM module also extracts the connection between `Manufacturability' and `Thickness' of the airfoil.}
    \label{fig:KG_retieval_manf}
\end{figure}

In a second example, the agent formulates a qualitative requirement for manufacturability. By querying the graph, the agent identified the conceptual link between `Design' and `Manufacturability' (Figure \ref{fig:KG_retieval_manf}), retrieving the principle that manufacturability is a key design consideration. The agent further synthesized context from related graph segments, such as the connection between `Manufacturability' and `Thickness' (Figure \ref{fig:KG_retieval_manf}), to reinforce its reasoning. This case highlights the system’s ability to not only retrieve discrete facts but also to incorporate broader contextual connections and inter dependencies captured within the knowledge graph, ensuring a more grounded set of initial requirements.

\begin{figure}[H]
    \centering
    \begin{textbox}{Knowledge Graph trace during requirement generation for manufacturability}
    \scriptsize
    "**Manufacturability**: The resulting airfoil geometry should be practical and feasible for manufacturing using conventional methods, avoiding features that would significantly increase production complexity or cost. \emph{(src:KG, Design, Manufacturability: The manufacturability of an airfoil design is an important consideration in the development process}. Reasoning: A design, however aerodynamically efficient, is not practical if it cannot be manufactured. The thickness distribution is a key aspect of manufacturability)."
  \end{textbox}
  \label{fig:functional_reqm}
\end{figure}

The RAG process requires an embedder for creating the vector database for the graph data against which the agents prompts are compared for extracting semantically similar text chunks. In this work, we use OpenAI's `text-embedding-3-small' model for creating this vectorized database for both our Knowledge Graphs. The results we obtain for knowledge retrieval is based on this version of embedding model, and variations in knowledge retrieval may exist when using other LLM models for embedding or graph generation.

\newpage
\section{Reasoning behind Systems Engineer agent's design review process}
\label{sec:agent_reasoning}
In this section, we provide an example showing the reasoning and steps used by the Systems Engineer agent while reviewing a design sample generated by the designer. The agent follows a structured reasoning workflow consisting of task interpretation, task decomposition, and action selection. The task assigned to the agent by the user is to use visual analysis along with geometric and aerodynamic data to evaluate the design's suitability against  technical requirements. The agent decomposes this task into the following action steps:
\begin{enumerate}
    \item Image analysis of airfoil profile plot
    \item Review of aerodynamic data provided by the Design Engineer
    \item Review of geometric parameters provided by the Design Engineer
    \item Evaluating design Quantities of Interest against Functional requirements
    \item Evaluating design against Non-Functional requirements
    \item Providing feedback and design improvement suggestions
\end{enumerate}
The excerpt below shows details of the reasoning used during the review process.

\begin{figure}[H]
    \centering
    \begin{textbox}{Excerpts from agent's reasoning during design review of sample ID-78}
    \scriptsize
    \textbf{Analyzing ID-78} \\
    
 \textbf{Final Result:}  In the context of airfoil design, this could represent the upper and lower surfaces of an airfoil profile, with `x' being the chordwise position and `y' being the vertical coordinate.\\
 
\textbf{Thought:} The user wants me to analyze an airfoil profile (ID-78) based on an image, aerodynamic data, and geometric parameters. I need to assess its suitability against provided functional and non-functional requirements, provide feedback, and determine its validity.\\

Here's a breakdown of the steps:
\begin{itemize}[leftmargin=0.2in]
    \vspace{-5pt}\item Analyze the image: The Vision Tool has described the image as a plot showing two curves, likely representing an airfoil profile. I need to infer the shape characteristics from this description.
    \vspace{-5pt}\item Review Aerodynamic Data: CD: 0.0084941151650085, CL: 1.142041730130238, CM: -0.3222857657483023
    \vspace{-5pt}\item Review Geometric Parameters: camber: 0.0762343152129009 (7.62\%), camber location: 0.7696591844688178 (76.97\% chord), thickness: 0.0459213124013521 (4.59\%)
    \vspace{-5pt}\item Compare with Functional Requirements:
        \begin{itemize}
            \vspace{-5pt} \item Airfoil Family: NACA 4-digit series geometry. (Need to check if the geometric parameters align with this, especially the thickness constraint).
            \vspace{-5pt} \item Design Objective: Maximize Cl/Cd. Calculate Cl/Cd.
            \vspace{-5pt} \item Operating Conditions: M=0.6, Re=$5\times10^6$, AoA=0 degrees. (Aerodynamic data is provided for these conditions).
            \vspace{-5pt} \item Geometric Parameters (NACA 4-digit Series Design Space)
                \begin{itemize}
                    \vspace{-2pt} \item Max Camber (m): 0\% <= m <= 9\%. Current: 7.62\%. (Valid)
                    \vspace{-2pt} \item Position of Max Camber (p): 0 <= p <= 9 (0\% to 90\% chord). Current: 76.97\% (p=7.69, which is 7 in NACA 4-digit). (Valid)
                    \vspace{-2pt} \item Max Thickness (t): 6\% <= t <= 20\%. Current: 4.59\%. (Invalid - too thin)
                \end{itemize}
            \vspace{-5pt} \item Performance Metrics: Cl, Cd, Cl/Cd. (Provided)
        \end{itemize}
\end{itemize}
\end{textbox}
  \label{fig:SE_reasoning_ID78_1}
\end{figure}

\begin{figure}[H]
    \centering
    \begin{textbox}{Excerpts from agent's reasoning during design review of sample ID-78 (cont'd)}
    \scriptsize
    \begin{itemize}[leftmargin=0.2in]
        \vspace{-5pt}\item Compare with Non-functional Requirements:
        \begin{itemize}
            \vspace{-2pt} \item Critical Mach Number: Aim > 0.6. (No direct data, but thin airfoils generally have higher critical Mach numbers, however, the thickness is *too* thin, which might lead to other issues).
            \vspace{-2pt} \item Boundary Layer Characteristics: Consider transition/separation at $Re=5\times10^6$ to minimize profile drag and ensure stable flow. (No direct data, but a very thin airfoil might have issues with boundary layer stability or structural integrity).
            \vspace{-2pt} \item Pressure Distribution**: Favorable pressure distribution for high lift/low drag at AoA=0, M=0.6. (No direct data, but the high Cl/Cd suggests a good pressure distribution).
            \vspace{-2pt} \item Computational Analysis Compatibility**: Assumed compatible.
            \vspace{-2pt} \item Manufacturability: Very thin airfoils can be challenging to manufacture and maintain structural integrity.
            \vspace{-2pt} \item Robustness**: Minor deviations should not lead to drastic performance degradation. (No direct data)
        \end{itemize}
    \vspace{-5pt}\item Check Cl value: Is Cl < 0.5? If yes, invalid. Current Cl = 1.142. (Valid, Cl > 0.5).
    \vspace{-5pt}\item Formulate Feedback: Combine image analysis, aerodynamic performance, and parameter values. Suggest improvements.
    \vspace{-5pt}\item Determine Validity: Based on all checks.
    \end{itemize}
    Let's start processing the information.

    \begin{itemize}[leftmargin=0.2in]
        \vspace{-5pt} \item Image Analysis (from Vision Tool): The image shows a very thin airfoil profile, consistent with the calculated thickness of ~4.6\%. The presence of two curves (blue and red) likely represents the upper and lower surfaces, and the overall shape appears to be cambered, aligning with the given camber value. The profile seems smooth, which is generally good for aerodynamic performance.        
        \vspace{-5pt} \item Geometric Parameters Analysis:
            \begin{itemize}
                \vspace{-2pt}\item Camber (m): 0.0762343152129009 -> ~7.6\% (within 0-9\% range). This corresponds to the first digit '7' in a NACA 4-digit series.
                \vspace{-2pt}\item Camber Location (p): 0.7696591844688178 -> ~77\% chord (within 0-90\% range). This corresponds to the second digit '7' in a NACA 4-digit series.
                \vspace{-2pt}\item Thickness (t): 0.0459213124013521 -> ~4.6\% (NOT within 6-20\% range). This corresponds to the last two digits '04' or '05' in a NACA 4-digit series.
            \end{itemize}
                The thickness constraint (6\% <= t <= 20\%) is violated. The current thickness is 4.59\%, which is too thin. This is a critical violation of the functional requirements.
        \vspace{-5pt} \item Aerodynamic Performance Analysis:
            \begin{itemize}
                \vspace{-2pt} \item CL: 1.142041730130238 (Greater than 0.5, so valid by this rule of thumb).
                \vspace{-2pt} \item  CD: 0.0084941151650085
                \vspace{-2pt} \item Cl/Cd = 1.142041730130238 / 0.0084941151650085 = 134.45
            \end{itemize}
            This is a very high lift-to-drag ratio, which is excellent for the design objective of maximizing Cl/Cd. The high lift coefficient at AoA=0 is expected due to the significant camber (7.6\%). The very low drag coefficient is also notable, possibly aided by the thin profile, but this thinness is a problem for the requirements.
        \vspace{-5pt} \item Suitability Assessment against Functional Requirements:
            \begin{itemize}
                \vspace{-2pt} \item  Airfoil Family: Assumed NACA 4-digit based on parameters, but the thickness constraint is violated.
                \vspace{-2pt} \item Design Objective: Maximize Cl/Cd. Cl/Cd = 134.45 is excellent.
                \vspace{-2pt} \item Operating Conditions: Data provided for M=0.6, $Re=5\times10^6$, AoA=0.
                \vspace{-2pt} \item Geometric Parameters: Camber (m=7.62\%): **Meets** (0-9\%), Camber Location (p=76.97\%): **Meets** (0-90\%), Thickness (t=4.59\%): **FAILS** (Required 6-20\%). This is a major issue.
                \vspace{-2pt} \item Performance Metrics: Cl, Cd, Cl/Cd are quantified. 
            \end{itemize}
    \end{itemize}    
  \end{textbox}
  \label{fig:SE_reasoning_ID78_2}
\end{figure}

\begin{figure}[H]
    \centering
    \begin{textbox}{Excerpts from agent's reasoning during design review of sample ID-78 (cont'd)}
    \scriptsize
        \begin{itemize} [leftmargin=0.2in]
            \vspace{-5pt} \item Suitability Assessment against Non-Functional Requirements:
            \begin{itemize}
                \vspace{-2pt} \item Aerodynamic Performance (Critical Mach Number): A very thin airfoil (like 4.6\%) would generally have a higher critical Mach number, which is good for M=0.6. However, the requirement specifies a minimum thickness of 6\% to manage drag at M=0.6, implying that *too* thin might also be problematic or not the intended design space.
                \vspace{-2pt} \item Boundary Layer Characteristics: A very thin airfoil might have challenges with boundary layer stability or transition at $Re=5\times10^6$, potentially leading to early separation if not carefully designed. The low drag coefficient suggests good boundary layer behavior, but the extreme thinness raises concerns.
                \vspace{-2pt} \item Pressure Distribution: The high Cl/Cd suggests a favorable pressure distribution.
                \vspace{-2pt} \item Computational Analysis Compatibility: Assumed compatible.
                \vspace{-2pt} \item Manufacturability: A 4.6\% thick airfoil can be very difficult to manufacture and maintain structural integrity, especially for larger structures or under significant loads. This is a significant concern.
                \vspace{-2pt} \item Robustness**: No direct data, but extremely thin airfoils can sometimes be less robust to off-design conditions.
            \end{itemize}

            \vspace{-5pt} \item Design Parameters and their effect:
                \begin{itemize}
                    \vspace{-2pt} \item Camber (0.0762): A non-zero camber is crucial for generating lift at AoA=0, as specified in the functional requirements. The current camber of 7.6\% is substantial and contributes significantly to the high lift coefficient (Cl=1.142).
                    \vspace{-2pt} \item Camber Location (0.7697): A camber location far aft (77\% chord) can influence the pitching moment and pressure distribution. For a NACA 4-digit, 'p' is usually between 0 and 9. A value of 7 (70\% chord) is common for aft-loaded airfoils. The current 76.97\% is quite far back, which might lead to a more negative pitching moment (CM = -0.322) and potentially affect stability.
                    \vspace{-2pt} \item Thickness (0.0459): The maximum thickness ratio is a critical parameter.
                    \vspace{-2pt} \item Impact on Drag: The requirement states "Excessive drag coefficient limits the maximum airfoil thickness ratio to values less than about 0.20". While the current thickness is well below 0.20, it is also below the *minimum* required thickness of 6\%. Very thin airfoils can have low profile drag, but they can also lead to structural issues and potentially adverse pressure gradients if not designed carefully.
                    \vspace{-2pt} \item Impact on Lift: Thickness also affects the maximum lift coefficient.
                    \vspace{-2pt} \item Impact on Critical Mach Number: Thinner airfoils generally have higher critical Mach numbers, which is beneficial for M=0.6.
                    \vspace{-2pt} \item Impact on Manufacturability: A thickness of 4.6\% is very thin and poses significant challenges for manufacturing and structural integrity.
                \end{itemize}
            \vspace{-5pt} \item Cl value check: Cl = 1.142041730130238, which is greater than 0.5. So, the design is *not* considered invalid based on this rule of thumb.
            \vspace{-5pt} \item Overall Validity: The design is **Invalid** primarily due to the violation of the minimum thickness requirement (4.59\% vs. required 6-20\%). While the aerodynamic performance (Cl/Cd) is excellent, this geometric constraint is fundamental. The very thin profile also raises concerns about manufacturability and structural integrity, which are non-functional requirements.
            \vspace{-5pt} \item Suggestions for Improvement: The primary suggestion is to increase the maximum thickness to be within the specified range of 6\% to 20\%. A thickness closer to the lower end of the acceptable range (e.g., 6-8\%) might still offer good aerodynamic performance at M=0.6 while meeting the structural and manufacturability considerations.
                \begin{itemize}
                    \vspace{-2pt} \item Increase Thickness: Aim for a thickness ratio (t) between 0.06 and 0.08. This will bring the design into compliance with the functional requirements and improve manufacturability and structural robustness.
                    \vspace{-2pt} \item Re-evaluate Camber Location: While 77\% chord for max camber is within the design space, it results in a significant negative pitching moment. Exploring a slightly more forward camber location (e.g., 40-60\% chord, corresponding to '4' or '5' in NACA 4-digit) might help in reducing the pitching moment while still generating sufficient lift at AoA=0. However, the primary objective is Cl/Cd maximization, and the current configuration achieves a very high Cl/Cd.
                    \vspace{-2pt} \item Monitor Aerodynamic Performance: After adjusting thickness, re-evaluate Cl, Cd, and Cl/Cd to ensure the design objective is still met. A slight increase in thickness might increase drag, but it should be manageable within the specified Mach number.
                \end{itemize}
        \end{itemize}
        \textbf{Feedback:} $\cdots$ \\
        \textbf{Valid:} False

    \end{textbox}
  \label{fig:SE_reasoning_ID78_3}
\end{figure}
\end{document}